\documentclass[]{ceurart}

\usepackage{listings}
\usepackage{url}
\usepackage{longtable}
\usepackage{booktabs}
\usepackage{array}
\usepackage{placeins}
\usepackage{float}
\usepackage{graphicx}
\begin{document}

\newcommand{\bs}[1]{{\color{teal} #1\textsuperscript{-bea}}}
\newcommand{\bscomment}[1]{\comment[id=bea]{#1}}

\newcommand{\ari}[1]{{\color{orange!70!black} #1\textsuperscript{-ari}}}
\newcommand{\aricomment}[1]{\comment[id=ari]{#1}}

\copyrightyear{2026}
\copyrightclause{Copyright for this paper by its authors.
  Use permitted under Creative Commons License Attribution 4.0
  International (CC BY 4.0).}

\conference{CLiC-it 2026: Twelfth Italian Conference on Computational Linguistics,
  September 14--16, 2026, Palermo, Italy}

\title{Do Language Models Know Their Slang? Queer Slang Understanding in User-Generated Content}

\author[1]{Arianna Denitto}[email=ariannadenitto6@gmail.com]\fnmark[1]
\author[2]{Beatrice Savoldi}[email=bsavoldi@fbk.eu]
\address[1]{University of Torino, Department of Humanities, Via S. Ottavio 20, 10124 Torino, Italy}
\address[2]{Fondazione Bruno Kessler, MT Unit, Via Sommarive 18, 38123, Povo (TN), Italy }

\fntext[1]{The work by Arianna Denitto was carried out during an internship at Fondazione Bruno Kessler between November 2025 and April 2026.}

\begin{abstract}
Despite its cultural relevance and diffusion, queer slang remains 
underrepresented in Natural Language Processing research. Towards addressing this gap, we introduce 
\textsc{Slang-Q}, a manually curated dataset of naturally user-generated English sentences 
paired with queer slang terms and reference definitions, built upon a 
newly constructed taxonomy of 118 queer terms. We use this resource to 
conduct a first exploratory evaluation of language models on their ability to 
understand and define queer slang under varying prompting conditions. 

\textsc{Slang-Q} is intended as a 
basis for studying how current models handle sensitive, community-specific language and whether they can provide accurate and reliable information about such forms of identity and linguistic expression.
\end{abstract}


\maketitle

\section{Introduction}

Large language models (LLMs) increasingly power conversational agents, 
mediate access to information, and underlie much of how people interact 
with text online \cite{chen2024combating, chatterji2025people, 
savoldi2026generativeaipracticesliteracy}. As these systems are deployed 
in diverse social contexts, their ability to handle the language of online 
communities becomes increasingly important \citep{basoah2025not}. Online, user-generated content poses a twofold challenge in this regard: community language 
evolves rapidly, making it difficult for models to keep up with linguistic 
change \citep{mei-etal-2024-slang}, and it is deeply tied to social 
identity. Slang, in particular, often serves as a marker of group 
membership and shared culture \citep{eble1996slang}, making its accurate 
interpretation a desirable property for models operating in user-facing 
settings.

\textit{Queer slang} exemplifies both of these challenges. It is a 
dynamic, creative, and community-specific sociolect that 
includes identity labels, reclaimed terms, spelling variations, shorthand 
forms, and idiomatic expressions whose meanings are often opaque outside 
their communities of use. Despite extensive research in queer linguistics 
\cite{kulick2000gay, motschenbacher2011taking} and community-led efforts 
to document and archive queer language \cite{cifor2023mediating}, queer 
slang remains underrepresented in Natural Language Processing (NLP) \cite{weber2026queer}. This has 
concrete consequences: queer terms are more likely to be flagged as hateful 
 \cite{dorn2024harmful}, and  underrepresentation contributes 
to cisheteronormativity in NLP systems more broadly 
\cite{vasquez2022heterocorpus}.

Existing work has addressed either generic slang handling 
\cite{mei-etal-2024-slang} or queer language primarily through the lens 
of bias and harm \cite{tint-2025-guardrails, weber2026queernlpcriticalsurvey, 
sosto-etal-2026-queergen}, but work at the intersection of the two remains 
limited. In this work, we focus on whether LLMs can provide accurate and 
reliable information about queer slang as it appears in user-generated 
content. Towards this goal, we introduce \textsc{Slang-Q},\footnote{The resource is made available at \url{https://github.com/ariadne28/Slang-Q}.} a resource comprising a taxonomy of 118 queer-related terms, from which we derive a  set of 1,204 manually annotated English user-generated sentences. With this, 
we evaluate the ability of proprietary and 
open-weight LLMs to understand and define queer slang under 
varying prompting conditions, probing the role of domain-specific framing 
and sentential context.

\section{Background}

\paragraph{Queer Slang} Following \citet{kulick2000gay} and recent work in NLP \cite{basoah2025not, tint-2025-guardrails}, we use \textit{queer slang} as a broad 
umbrella term encompassing the words, phrases, and expressions associated 
with LGBTQIA+ communities--- including identity labels, reclaimed terms, 
and community-specific idioms.
Queer slang has long been characterized by lexical creativity, including expressions that circulate within specific communities of practice and whose meanings often depend on shared cultural knowledge and context. Historical examples, such as the \textit{Polari}\footnote{\url{https://kittensyzygy.github.io/polari-dictionary/home.html}} expression «\textit{friend of Dorothy}»\footnote{A gay slang term for a gay man, and more broadly, any Queer-identified person. A euphemism that originates as far back as World War II in the United States, when homosexual acts were illegal – thus, a way of asking about someone’s sexual orientation without being explicit. \cite{librarylgbt}}, show how queer slang originally may have served as a means for mutual recognition and protection, while contemporary digital environments have accelerated its circulation and transformation. Queer slang also often overlaps with other linguistic practices, such as African American Vernacular English (AAVE), making it difficult to isolate from other minoritized dialects \cite{tint-2025-guardrails,veloso2025slayingqueerlanguageprocessing}.
This poses a challenge for language technologies. Recent work has shown that LLMs can effectively grasp the complexities of evolving slang language if supported by robust methodologies \cite{mei-etal-2024-slang}, although such models are widely known to still exhibit cultural biases and struggle with non-standard language, especially when coming from marginalized contexts such as the queer community \cite{tint-2025-guardrails}. This issue is particularly relevant for user-generated content, where informal spellings, neologisms, abbreviations, and shifting senses are common. In our setting, the problem is compounded by the fact that many queer-related terms are polysemous: words such as \textit{drag}, \textit{bear}, or \textit{tea} may occur with queer-specific meanings, but also with more common, unrelated everyday senses.

\paragraph{Slang and Queer Language in NLP} A growing body of work has examined queer-related language in NLP. Recent surveys show that queer NLP has expanded in the last few years, but much of the field remains focused on exposing bias and other issues rather than developing new resources or solutions \cite{weber2026queernlpcriticalsurvey}. Common objects of study include neopronouns and gender identity \cite{piergentili-etal-2024-enhancing,ungless-etal-2025-amplifying}, hate speech, slurs and stereotypes, especially on platforms like social media \cite{locatelli-etal-2023-cross,pofcher-etal-2025-hope}. Most of these studies usually highlight how some queer-coded language may be mistakenly classified by models as slurs, harmful or negative expressions even when they carry a neutral, positive, or reclaimed stance \cite{weber2026queernlpcriticalsurvey, tint-2025-guardrails}. Existing research has investigated how NLP systems represent and process queer-related language, including constrained queer narratives in LLM-generated text \cite{ghosal2025unequalvoicesllmsconstruct}, gender and sexuality norms reflected in sentence completion \cite{sosto-etal-2026-queergen}, representation of queer-related words in lexicons \cite{ramesh-etal-2022-revisiting}, and misgendering in model outputs \cite{subramonian2025agreedisagreemetaevaluationllm}.

Recent work that is close to our research \citep{tint-2025-guardrails} evaluates how queer slang in prompts affects the emotional content of LLM outputs, finding that queer slang is associated with more negative emotional responses than comparable heteronormative language. The study therefore highlights persistent biases in how LLMs process community-specific queer language, even when overtly harmful responses are mitigated. Moreover, \textsc{SLAyiNG} \cite{veloso2025slayingqueerlanguageprocessing} has recently been introduced as an ongoing manually annotated dataset explicitly focused on queer slang, framing its processing as a sense disambiguation task in noisy real-world examples and highlighting the need for expert and community-driven annotation. Such a resource, however, is not publicly available.

Overall, while these studies have advanced our understanding of how queer language and identities interact with NLP systems, they have  primarily treated queer slang or identities as a variable affecting model behaviour--- examining bias, harm, or output sentiment---rather than as an 
object of knowledge in its own right. In this work, we shift the focus 
to what models actually \textit{know} about queer slang: whether they can accurately understand community-specific expressions as they appear in 
user-generated content. To support this investigation, we also provide \textsc{Slang-Q},  
a new annotated resource for the community. Together, these 
contributions bring our evaluation closer to settings 
in which LLMs increasingly operate---e.g. as sources of information, 
mediators of online interaction, and tools for text understanding.

\section{Data Creation}
\label{sec:data}

\textsc{Slang-Q} is a manually curated evaluation set consisting of 1,024 English sentences containing queer-related terms, where each entry pairs a naturally occurring sentence with its annotated matched term and a reference definition.
To construct the dataset, we used  Urban Dictionary data,\footnote{\url{https://www.urbandictionary.com/}} previously extracted and released within the generic \textsc{SLANG} benchmark  \citep{mei-etal-2024-slang}.

This source is suitable for our purposes because it contains naturally occurring examples of informal and slang language generated by users, and is publicly available under a permissive license.\footnote{\url{https://github.com/Meirtz/FocusOnSlang-Toolbox/blob/main/LICENSE}} At the same time, the Urban Dictionary data is noisy and not queer-specific: entries may contain irrelevant senses, mismatched examples, or vulgar and harmful content.

To construct \textsc{Slang-Q}, we therefore combine automatic extraction with manual filtering. Concretely, our pipeline proceeds in four steps: 
(\textbf{i.})~compilation of a queer-related term inventory, 
(\textbf{ii.})~term-based extraction of candidate sentences from the 
source data, and (\textbf{iii.})~manual annotation for semantic relevance 
and harmfulness. Each step is described in detail below.

\subsection{Term List Creation and Taxonomy}
To filter the source dataset, we compile a list of queer-related terms and use it to extract candidate user-generated examples containing these expressions. Each term is paired with its definition and categorical labels. 

\subsubsection{Queer Terms Compilation and Definitions}

Since queer slang is heterogeneous and rapidly changing, we draw on different types of sources (e.g. research papers or online archives) to balance lexical variety and coverage.

First, we retain 46 out of 57 
terms from the list provided by \citet{tint-2025-guardrails}, excluding vulgar or non-reclaimed lexicon.  We further include 24 terms from the queer lexicon of 
\citet{ramesh-etal-2022-revisiting}, which was specifically designed 
to audit the coverage of queer minorities in lexical resources for 
online abuse and hate speech detection. To expand the list beyond these resources, we add terms from publicly available lexical and community-oriented sources. We extract 5 terms from Wikipedia's \textbf{LGBTQ slang} category,\footnote{\url{https://en.wikipedia.org/wiki/Category:LGBTQ_slang}} which provides a broad overview of terms commonly recognized as queer-related slang. We include 26 terms from \textbf{Lexicon Library.LGBT} \cite{librarylgbt}.

Finally, we add 4 terms from \textbf{Wiktionary},\footnote{\url{https://en.wiktionary.org/wiki/}} where definitions and usage notes are available for some informal or emerging lexical items not covered by the other sources. After a first annotation round, 13 further terms are added because they emerge directly from the user-generated examples and prove relevant to the scope of the dataset. The final inventory contains 118 unique terms. 
For each term, we collect a gold definition to be used as a reference in the evaluation (see \S\ref{sec:eval}). When a term is added from a lexical source, such as Lexicon Library.LGBT, Wiktionary, or Wikipedia's LGBTQ slang category, we use the definition provided by that source. These same sources are also prioritized for terms collected from previous NLP work \citep{tint-2025-guardrails,ramesh-etal-2022-revisiting}, since such papers provide term inventories but not definitions. When no definition is available from the primary sources, we rely on complementary archives, e.g. Chew Inclusive's LGBTQIA+ terminology glossaries.\footnote{\url{https://itg.nls.uk/wiki/Introduction}}
\enlargethispage{1\baselineskip}

\subsection{Queer Terms Taxonomy}

Inspired by the \textbf{Slang section} of \textbf{Lexicon Library.LGBT},\footnote{As of July 29, 2026, Lexicon Library.LGBT was discontinued (per correspondence with the site administrators) and is no longer available at its original URL (\texttt{https://lexicon.library.lgbt}). Pages cited here remain accessible via the  Internet Archive’s Wayback Machine.}
we organize our lexical resource around a taxonomy in which each term is enriched with one or more broad categories.

The taxonomy distinguishes between terms that primarily denote identities, terms used as queer slang, and intersectional terms that are often used within queer communities but also fall under the definitions of internet slang, AAVE, or other linguistic and cultural contexts, such as fandom.\footnote{Term referring to a subculture that gathers fans united by a common interest.} Rather than enforcing a strict boundary in such cases, we used the \textit{Intersectional} label to mark expressions whose relevance to queer communities coexists with wider circulation across other online or minoritized language practices. All choices, including decisions about term meanings and potentially harmful usage, were discussed between the two authors and also guided by the categorizations and discussions found in the resources consulted for this work.\footnote{\url{https://lexicon.library.lgbt/about/}} The taxonomy is organized on two levels to capture both the semantic domain of each item and the linguistic mechanism through which it functions. As described above, broad categories describe the area of queer-related terminology to which a term belongs; the subcategories, instead, capture linguistic phenomena that cut across these domains. 

In practice, each term is assigned to at least one broad main category, whereas subcategories are optional and only apply when a term displays a specific linguistic phenomenon, such as shorthand formation or spelling variation.
The broad categories are: \textit{Identity}, \textit{Slang}, and \textit{Intersectional}; the subcategories include \textit{Reclaimed}, \textit{Shorthand}, \textit{Idiomatic expression}, \textit{Pronoun}, and \textit{Spelling variation}. The taxonomy is summarized in Table~\ref{tab:categories}, whereas the complete list of terms associated with their (sub)categories is reported in Appendix~\ref{app:term-list}.

\begin{table}[!tbp]
\centering
\begin{tabular}{p{0.18\linewidth} p{0.56\linewidth} p{0.18\linewidth}}
\toprule
\textbf{Label} & \textbf{Definition} & \textbf{Example terms} \\
\midrule
\multicolumn{3}{l}{\textsc{Categories}} \\
\cmidrule(lr){1-3}
\addlinespace[0.2em]
Identity & Terms referring to identity, gender, and sexual orientation. & \textit{sapphic, enby} \\
Slang & Slang specifically used within the queer community. & \textit{beard, closeted} \\
Intersectional & Slang used within the queer community as well as in other communities, such as AAVE or fandom. & \textit{slay, tea} \\
\addlinespace[0.4em]
\midrule
\multicolumn{3}{l}{\textsc{Subcategories}} \\
\cmidrule(lr){1-3}
\addlinespace[0.2em]
Reclaimed & Terms originally used in a derogatory way and later reappropriated within the community with an empowering nuance. & \textit{queer, butch} \\
Shorthand & Acronyms, initialisms, abbreviations, and other shortened forms. & \textit{ace, cishet} \\
Idiomatic 

expression & Figurative expressions used in queer slang. & \textit{coming out, slap} \\
Pronoun & Neopronouns used within the queer community. & \textit{ze, xe} \\
Spelling variation & Terms whose spelling is altered in slang usage while preserving meaning. & \textit{kween, yas} \\
\bottomrule
\end{tabular}
\caption{Categories and subcategories for queer-related terms.}
\label{tab:categories}
\end{table}

\subsection{Extraction and Annotation}
\label{sec:stats}
We apply our inventory of 118 terms to the source data via string 
matching, extracting all sentences containing at least one matched 
term. The source corpus consists of 185,795 Urban Dictionary entries 
collected in 2024 \cite{mei-etal-2024-slang}. The automatic extraction yields 3,184 initial
sentences.

Manual inspection of the extracted sentences reveals that several 
terms frequently occur with meanings unrelated to queer slang: 
\textit{bear}, \textit{tea}, or \textit{drag}, for instance, are 
common homonyms whose dominant sense in the data is often non-queer. 
We therefore proceed with a two-level manual annotation step. For 
each sentence, we assess: (\textbf{i.})~whether the matched 
term is used with its intended queer slang meaning, and 
(\textbf{ii.})~whether the example contains harmful or vulgar content, 
which we discard in order to retain only instances reflecting positive 
or neutral community usage.

Out of the 3,184 extracted sentences, 1,833 (57.57\%) were discarded 
as semantically unrelated to the intended meaning, and 327 (10.27\%) 
were removed due to harmful or vulgar content in the context of the sentence.

Annotation was carried out by the first 
author---who has a background in digital humanities and expertise in online community language, including fandom slang and queer internet vernacular. Fewer ambiguous cases were discussed with the second 
author. Overall, a conservative approach was adopted whereby any entry was deemed harmful or vulgar by either of the authors. The annotation explicitly accounted for reclaimed usage: examples were retained when sentence structure and tone clearly indicated positive or neutral in-group use. However, because Urban Dictionary provides no reliable author metadata, genuinely ambiguous or hostile cases were excluded as a safety measure.

\subsection{Dataset Statistics}
The final dataset consists of 1,024 unique entries, each pairing a 
sentence with its matched queer slang term and a reference definition. 
Because the 118-term inventory was built independently of the source corpus, not every term was guaranteed to occur in the data. In practice, 77 of the 118 terms (65.25\%) appeared in valid examples and were retained for evaluation; the remaining terms, while absent from the dataset, remain part of our published taxonomy.

An example entry is 
shown in Table~\ref{tab:main-sample-annotations}, and statistics by taxonomy category 
are reported in Table~\ref{tab:category-coverage}.

\begin{table}[!htbp]

\centering

\scriptsize

\begin{tabular}{p{0.22\linewidth}p{0.70\linewidth}}

\hline

\textbf{Field} & \textbf{Value} \\

\hline

Matched term(s) & closeted \\

\hline

Urban Dictionary example & 
\begin{minipage}[t]{\linewidth}
\itshape
Person 1: I guess the biggest thing people don't know about me is that I'm a deeply closeted gay man

person 2: You're a gay guy?

Person 1: I'm not gay! I'm deeply closeted!
\end{minipage}
\\

\hline

Category & Identity, Slang \\

\hline

Subcategory & Idiomatic expression \\

\hline

Definition & 
\begin{minipage}[t]{\linewidth}
Individual who is not straight or cisgender but is currently not open to everybody (or anybody) about their identity. There are a number of reasons for being “in the closet” (for example, a fear of rejection or one’s safety, disapproval from friends/family, or discrimination).
\end{minipage}

\\

\hline

\end{tabular}

\caption{Example of a final \textbf{Slang-Q} dataset entry.}

\label{tab:main-sample-annotations}

\end{table}

Two aspects of the dataset structure are worth noting. First, some 
terms belong to more than one category (e.g., \textit{bi} is annotated 
as \textit{Identity} only, while \textit{friend of dorothy} is annotated as both 
\textit{Identity} and \textit{Slang}), so category-level counts are non-mutually exclusive 
and sum to more than 1,024. Second, sentences containing more than one 
matched term are treated as separate entries, one per term, each paired 
with its corresponding definition.

At the occurrence level, slang terms are the most frequent category, 
followed by intersectional and identity-related terms 
(Table~\ref{tab:category-coverage}). This distribution is partly driven 
by the high frequency of a small number of terms, such as \textit{goat}, 
\textit{slay}, \textit{bae}, \textit{trans}, and \textit{gurl}. 
Additional statistics, including the distribution of the most frequent 
matched terms and subcategory-level coverage, are reported in 
Appendix~\ref{app:stats}.

\begin{table}[!htbp]
\centering
\small
\begin{tabular}{lrrr}
\toprule
\textbf{Category} & \textbf{Terms} & \textbf{Matched terms} & \textbf{Sentences} \\
\midrule
Slang & 65 & 39 & 266 \\
Identity & 65 & 40 & 294 \\
Intersectional & 29 & 22 & 612 \\
\bottomrule
\end{tabular}
\caption{Term and sentence counts in \textsc{Slang-Q} by taxonomy 
category. \textit{Terms} refers to the number of terms assigned to 
each category in the original inventory of 118 items; 
\textit{Matched terms} refers to those that occur at least once in 
the final dataset. Categories are non-mutually exclusive.}
\label{tab:category-coverage}
\end{table}

\section{Experimental Settings}
\label{sec:experiments}

\subsection{Models}
We selected state-of-the-art instruction-following LLMs available at the time of our experiments, covering both proprietary and open-weight systems. 
 Due to computational resource constraints, we limited our evaluation to four models. 
As a commercial model, we included Claude Sonnet 4.6,\footnote{\url{https://www.anthropic.com/news/claude-sonnet-4-6}} accessed via the Anthropic API.\footnote{\url{https://www.anthropic.com/api}} For open-weight models, we included Qwen3 32B \cite{qwen3technicalreport},\footnote{\url{https://huggingface.co/Qwen/Qwen3-32B}} LLaMA 4 Scout \cite{arxiv2026llama4herdarchitecture},\footnote{\url{https://huggingface.co/meta-llama/Llama-4-Scout-17B-16E}} and LLaMA 3.3 70B \cite{grattafiori2024llama3herdmodels}.\footnote{\url{https://huggingface.co/meta-llama/Llama-3.3-70B-Instruct}}, using the Groq API\footnote{\url{https://console.groq.com}. We used default parameters.} for inference. 
Overall, this selection allowed us to compare models across different families as well as more recent against older releases. 

\begin{table*}[t]
\centering
\footnotesize
\begin{tabular}{llp{10cm}}
\toprule
\textbf{Condition} & \textbf{Input} & \textbf{Prompt} \\
\midrule
Term baseline & term & 
\textit{You are a language expert. Given a term below, provide a clear and concise definition in English. Write 1--3 sentences. Avoid unnecessary elaboration. Avoid including usage examples unless essential for meaning. Only list multiple meanings when both are used frequently in similar contexts.} \newline \texttt{Term: \{term\}} \\
\addlinespace
Term slang & term & 
\textit{You are a queer internet slang expert. Given the queer slang term below, provide a clear and concise definition in English. Write 1--3 sentences. Avoid unnecessary elaboration. Avoid including usage examples unless essential for meaning. Only list multiple meanings when both are used frequently in queer contexts.} \newline \texttt{Term: \{term\}} \\
\addlinespace
Context baseline & term + example & 
\textit{You are a language expert. Given a term and an example sentence, provide a clear and concise definition in English. Use the example sentence to infer the intended meaning. Write 1--3 sentences. Avoid unnecessary elaboration. Avoid including usage examples unless essential for meaning. Only list multiple meanings when both are used frequently in similar contexts.} \newline \texttt{Term: \{term\}} \newline \texttt{Example: \{example\}} \\
\addlinespace
Context slang & term + example & 
\textit{You are a queer internet slang expert. Given the queer slang term and an example sentence, provide a clear and concise definition in English. Use the example sentence to infer the intended meaning. Write 1--3 sentences. Avoid unnecessary elaboration. Avoid including usage examples unless essential for meaning. Only list multiple meanings when both are used frequently in queer contexts.} \newline \texttt{Term: \{term\}} \newline \texttt{Example: \{example\}} \\
\bottomrule
\end{tabular}
\caption{Overview of the four experimental conditions and their corresponding prompts.}
\label{tab:conditions}
\end{table*}

\subsection{Experimental Conditions}
    \label{sec:conditions}

To assess whether models can support queer slang, 
we frame our evaluation as a definition generation task, in which models 
are prompted to provide a concise definition of a given term. 
We evaluated all models under four conditions, resulting from the combination of two dimensions: the amount of context provided to the model, and whether the prompt was framed generically or was slang-informed.

The first dimension contrasted a \textbf{term}-only condition, in which the model was given only the target term, with a \textbf{context} condition, in which the term was accompanied by the example sentence from our dataset. 
This served to explore the role of  
context in disambiguating terms.

The second dimension, instead, contrasted a \textbf{baseline} prompt, in which the model was instructed to act as a generic language expert, with a \textbf{slang}-informed prompt, in which the model was explicitly cast as a queer internet slang expert. The 
baseline vs.\ slang-informed dimension assessed how models behaved under 
a default, naturalistic prompting condition---as a user might casually 
query a model---compared to a domain-specific framing that explicitly 
oriented the model toward queer slang.

As shown in Table \ref{tab:conditions}, the four resulting conditions are: \textit{(i)}~term baseline, 
\textit{(ii)}~term slang-informed, (iii)~context baseline, and 
(iv)~context slang-informed. In all conditions, models were instructed to generate a definition 1-3 sentences long---reflecting the length of our human gold definitions (see \S\ref{sec:eval})---and received a one-shot 
exemplar to ensure format adherence; the example was a generic term definition 
in the baseline conditions and a queer slang definition in the slang-informed 
conditions, such that the shot itself also reflected the intended framing.

\subsection{Evaluation Methods}
\label{sec:eval}

As described in Section~\ref{sec:conditions}, we framed our evaluation 
as a definition generation task, in which models were prompted to 
produce a definition of a given term. To evaluate model outputs 
automatically, we compared generated definitions against gold reference 
definitions using two complementary metrics: \textbf{ROUGE-L} 
\citep{lin-2004-rouge}\footnote{\url{https://github.com/google-research/google-research/tree/master/rouge}} 
and \textbf{BERTScore} \citep{zhang2020bertscoreevaluatingtextgeneration}.\footnote{\url{https://huggingface.co/papers/1904.09675}} 
ROUGE-L measures the longest common subsequence overlap between a 
generated and a reference definition, capturing lexical and structural 
similarity. BERTScore computes similarity via contextual embeddings, 
capturing semantic closeness beyond surface form. Together, the two 
metrics allowed us to evaluate both surface-level overlap and semantic closeness.

Before applying the metrics to model outputs, we conducted a validation 
step to assess their behaviour under length variation. Starting from 
all the gold definitions in our taxonomy (avg. length 21.2 words), 
 we 
used \textbf{GPT-5.5}\footnote{\url{https://developers.openai.com/api/docs/models/gpt-5.5}} 
to generate, for each of the 118 terms, three meaning-preserving 
alternatives of decreasing length---averaging 22.3, 15.6, and 11.4 words 
respectively---and manually reviewed them to ensure soundness. 
As expected, ROUGE-L decreased as definitions became shorter, 
from 0.627 for the longest alternatives to 0.438 and 0.386 for the 
shorter ones, reflecting its sensitivity to lexical overlap. BERTScore 
followed the same trend but with a smaller drop, from 0.924 to 0.905 
and 0.896, confirming its greater robustness to surface variation. This 
validated our use of both metrics: ROUGE-L as a stricter lexical 
measure and BERTScore as a more semantically oriented one.

In the \textbf{main experiments}, we use a multi-reference approach to ensure robustness. Each output was compared against three reference definitions: the original human-authored taxonomy definition, and the two shorter GPT-5.5-generated alternatives, which were manually validated for meaning preservation. The longest alternative 
was excluded as it closely mirrored the length of the original definition and would 
therefore be redundant. Final scores were reported as the mean across 
the three references. To avoid bias introduced by terms that occur 
frequently in the dataset, we reported term-level aggregated scores: 
scores were first averaged across all sentences for a given term, and 
then averaged across terms, so that each term contributed equally 
to the final evaluation regardless of its frequency.

\subsection{Preliminaries}
\label{sec:contamination}

Online user-generated content, such as that represented in our dataset, 
risks being integrated into model training data.
As a preliminary verification, we thus employed the Data Contamination Quiz (DCQ) framework 
\cite{golchin-surdeanu-2025-data}, a black-box method that detects and quantifies verbatim 
contamination without requiring access to model weights or training data. The method frames contamination detection as a series of multiple-choice questions in which the model must identify an original dataset instance among word-level perturbations of it.

Results indicated absent to low levels of verbatim memorisation across 
open-weight models: LLaMA 4 Scout [10.75\%--17.00\%], LLaMA 3.3 70B 
[10.53\%--15.00\%], and Qwen3-32B [0.00\%--0.00\%]. Claude Sonnet 4.6, 
instead, showed a markedly higher contamination range [57.30\%--62.00\%], 
suggesting greater exposure to user-generated content of this kind during 
training. Importantly, verbatim memorisation of an example sentence does 
not entail the ability to solve the definition generation task, as our 
evaluation did not rely on Urban Dictionary definitions as ground truth 
labels. These results should therefore be read as a preliminary indication 
of how much user-generated content of this kind is represented in each 
model's training data, and may help interpret performance differences in 
the main experiments. Full details on the approach and procedure are given 
in Appendix~\ref{app:data-contamination}.

\section{Results}
\label{sec:results}

\begin{table*}[htp!]
\small
\centering
\label{tab:term-eval}
\setlength{\tabcolsep}{4pt}
\begin{tabular}{@{}l|cccc|cccc@{}}
\toprule
& \multicolumn{4}{c|}{ROUGE-L} & \multicolumn{4}{c}{BERTScore F1} \\
 & Term & Term-S & Context & Context-S & Term & Term-S & Context & Context-S \\
\cmidrule(lr){2-5} \cmidrule(lr){6-9}

Human & \multicolumn{4}{c|}{\cellcolor[RGB]{80, 150, 230} 0.48 \scriptsize{± 0.13}} & \multicolumn{4}{c}{\cellcolor[RGB]{80, 150, 230} 0.91 \scriptsize{± 0.03}} \\

\midrule
claude 4.6 & \cellcolor[RGB]{228,238,250} 0.16 \scriptsize{± 0.07} & \cellcolor[RGB]{196,216,240} 0.18 \scriptsize{± 0.06} & \cellcolor[RGB]{221,233,247} 0.17 \scriptsize{± 0.06} & \cellcolor[RGB]{196,216,240} 0.18 \scriptsize{± 0.06} & \cellcolor[RGB]{228,238,250} 0.85 \scriptsize{± 0.02} & \cellcolor[RGB]{196,216,240} 0.86 \scriptsize{± 0.01} & \cellcolor[RGB]{228,238,250} 0.85 \scriptsize{± 0.01} & \cellcolor[RGB]{196,216,240} 0.86 \scriptsize{± 0.01} \\
llama3.3 70B & \cellcolor[RGB]{233,242,251} 0.16 \scriptsize{± 0.06} & \cellcolor[RGB]{221,233,247} 0.17 \scriptsize{± 0.05} & \cellcolor[RGB]{232,241,251} 0.16 \scriptsize{± 0.05} & \cellcolor[RGB]{221,233,247} 0.17 \scriptsize{± 0.05} & \cellcolor[RGB]{228,238,249} 0.85 \scriptsize{± 0.02} & \cellcolor[RGB]{196,216,240} 0.86 \scriptsize{± 0.02} & \cellcolor[RGB]{196,216,240} 0.86 \scriptsize{± 0.01} & \cellcolor[RGB]{196,216,240} 0.86 \scriptsize{± 0.01} \\
llama4 & \cellcolor[RGB]{245,250,255} 0.14 \scriptsize{± 0.06} & \cellcolor[RGB]{221,233,247} 0.17 \scriptsize{± 0.05} & \cellcolor[RGB]{230,239,250} 0.16 \scriptsize{± 0.05} & \cellcolor[RGB]{232,241,251} 0.16 \scriptsize{± 0.05} & \cellcolor[RGB]{245,250,255} 0.85 \scriptsize{± 0.02} & \cellcolor[RGB]{196,216,240} 0.86 \scriptsize{± 0.02} & \cellcolor[RGB]{196,216,240} 0.86 \scriptsize{± 0.02} & \cellcolor[RGB]{196,216,240} 0.86 \scriptsize{± 0.01} \\
qwen3 32B & \cellcolor[RGB]{227,238,249} 0.16 \scriptsize{± 0.07} & \cellcolor[RGB]{221,233,247} 0.17 \scriptsize{± 0.07} & \cellcolor[RGB]{196,216,240} 0.18 \scriptsize{± 0.07} & \cellcolor[RGB]{221,233,247} 0.17 \scriptsize{± 0.06} & \cellcolor[RGB]{237,244,252} 0.85 \scriptsize{± 0.02} & \cellcolor[RGB]{196,216,240} 0.86 \scriptsize{± 0.02} & \cellcolor[RGB]{196,216,240} 0.86 \scriptsize{± 0.02} & \cellcolor[RGB]{196,216,240} 0.86 \scriptsize{± 0.01} \\
\bottomrule
\end{tabular}
\caption{Mean and standard deviation of BERTScore F1 and ROUGE-L across models, with each term across sentences contributing equally. Human refers to the mean score of the gold reference definition against the two  alternative definitions.}
\end{table*}

Table~\ref{tab:term-eval} reports mean ROUGE-L and BERTScore across 
models and conditions. As a general observation, all models score 
below the human upper bound across both metrics, suggesting that 
definition generation for queer slang is non-trivial. 
At the same time, scores are relatively high and consistent, 
indicating that models often produce semantically accurate outputs. In particular, \textbf{across metrics}, BERTScore showes little variation across models and conditions, 
with all systems scoring in a narrow range, close to the human 
upper bound. ROUGE-L reveals more variation and a larger gap with 
respect to human scores, likely due to its higher sensitivity to 
lexical overlap. 

The most consistent trend is the effect of \textbf{prompting condition}. 
The Term (Base) condition---where the model is given only the 
target term with no indication that it refers to queer slang--- 
yields the lowest scores across all models and for both metrics. We attribute 
this to the inherent polysemy of several terms in the dataset: 
expressions such as \textit{bear}, \textit{tea}, or \textit{drag} 
carry everyday meanings that may interfere with queer-specific 
interpretations when no disambiguating signal is provided. 
Scores improve when either a slang-informed framing is used 
(Term-S) or a sentence-level example is included (Context 
conditions), suggesting that both explicit domain orientation 
and contextual grounding help models converge on the intended 
queer meaning. 
\textbf{Across models}, differences are small or absent, 
with no single model standing out substantially. This suggests 
that the models evaluated have broadly similar knowledge of 
queer slang as it appears in online user-generated content, 
despite their likely different exposure to Urban Dictionary data (see \S~\ref{sec:contamination}).

\begin{figure}[htp!]
    \centering
    \includegraphics[width=1\linewidth]{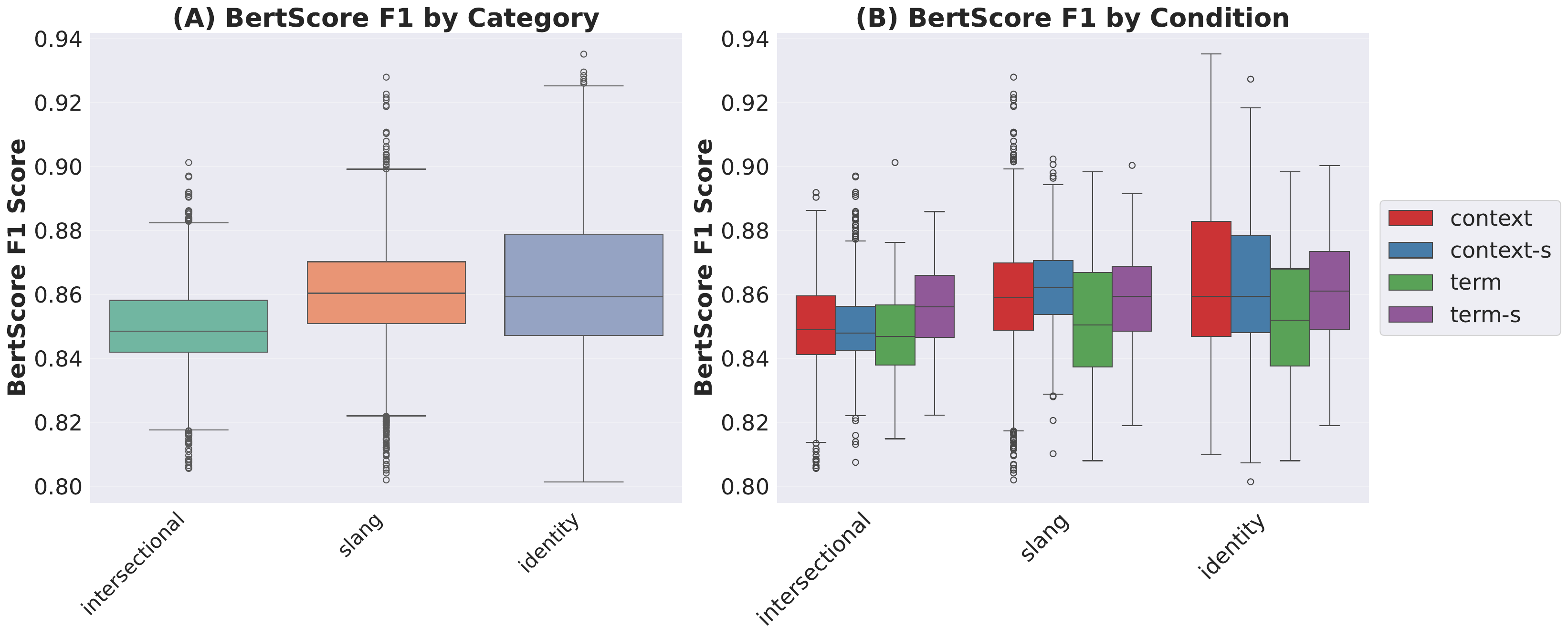}
\caption{BERTScore F1 scores by taxonomy category, aggregated across 
all models and conditions (A) and broken down by prompting condition (B).}    \label{fig:bert_category}
\end{figure}

\begin{figure} [htp!]
    \centering
    \includegraphics[width=1\linewidth]{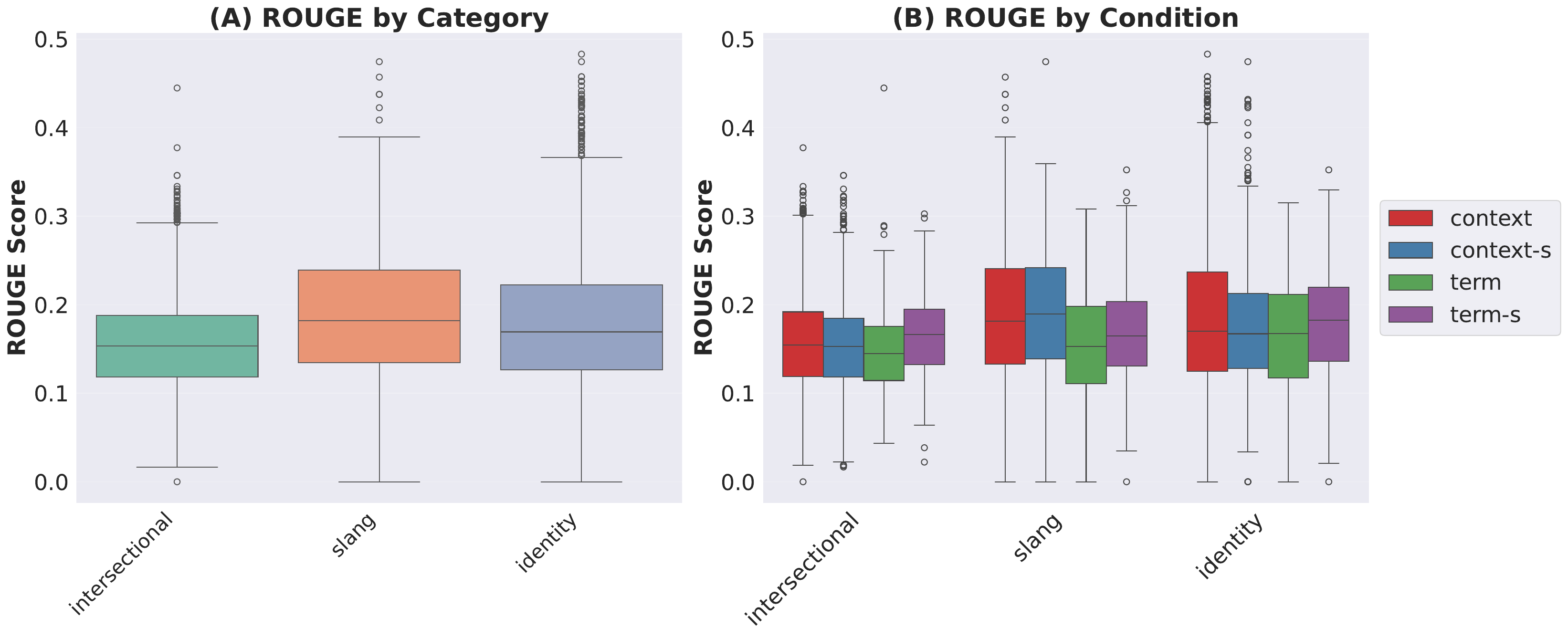}
\caption{ROUGE scores by taxonomy category, aggregated across 
all models and conditions (A) and broken down by prompting condition (B).}    \label{fig:rouge_category}
\end{figure}

\paragraph{Category Results}
Figures~\ref{fig:rouge_category} and~\ref{fig:bert_category} break 
down ROUGE-L and BERTScore by taxonomy category, respectively. 
Looking at the aggregated results (panel A in both figures), 
intersectional terms consistently score lowest across both metrics, 
which we attribute to their inherently ambiguous nature: these terms 
circulate across multiple communities---such as queer spaces and 
AAVE---and their intended queer-specific meaning may therefore be 
more nuanced. Slang and identity terms score comparably higher 
and showed greater variance, reflecting the internal heterogeneity of 
these larger categories. The same ranking holds under BERTScore, though with 
smaller differences, consistent with its greater robustness to surface 
variation. The condition breakdown (panel B) confirms that the base Term-only
condition is the most challenging across all three categories. 
Notably, slang terms show the greatest benefit from both slang-informed 
framing and contextual grounding.
This suggests 
that for slang terms in particular---whose meanings are often 
idiomatic--- additional framing helps models 
converge on the intended interpretation. 

\paragraph{Manual Evaluation}
We complement our automatic evaluation with a manual error analysis on a subset of models and conditions. Namely, we focus on the two most contrasting conditions---Term baseline and slang-informed Context---for \textbf{Claude Sonnet 4.6} and \textbf{Llama 4 Scout}. Indeed, the two models show the clearest performance contrast in ROUGE-L: Claude scores higher in both conditions, while Llama 4 records the lowest Term-baseline ROUGE-L among all evaluated models, despite comparatively close BERTScore results. This makes the pair well suited to test whether similar semantic similarity conceals qualitative differences. For each model and across both conditions, we manually inspect 50 outputs sampled across our main term categories, for a total of 200 outputs. Definitions are labeled as \textit{Correct} when they convey the intended meaning, \textit{Partially correct} when they identify the relevant slang or queer sense but omit or distort a central component, and \textit{Incorrect} when they select an unrelated sense, hallucinate a meaning, or fail to define the term.

Manual inspection confirms a strong effect of prompting condition. In the \textbf{Term baseline}, Claude produces 70\% correct, 4\% partially correct, and 26\% incorrect definitions, while Llama 4 obtains 64\%, 8\%, and 28\%, respectively. With \textbf{slang-informed context}, Claude reaches 100\% correct outputs, whereas Llama 4 produces 90\% correct and 10\% partially correct definitions.

The qualitative analysis reveals several recurring patterns. Without context, models tend to default to the dominant general-language meaning of polysemous terms, as in the case of \textit{bear}. However, some queer meanings appear sufficiently conventionalized to be retrieved without additional framing, as happens with both models and the expression \textit{coming out}. The models instead struggle more with novel, uncommon, or morphologically derived slang. They may state that a term is unknown, treat it as a misspelling, or hallucinate information. In most cases, sentence context provides enough evidence to recover the intended meaning. For example, both models fail to define \textit{lesbotorium} in isolation but infer its meaning correctly from context.

Llama 4 also frequently prefaces an answer by stating that it cannot identify a term before using explicit reasoning to derive a correct or partially correct meaning. Such responses reveal at least partial knowledge, but their disclaimers and extended reasoning introduce material that is absent from the reference definition. More generally, some outputs over-explain the accompanying context or rely too heavily on it instead of providing a concise definition. Others recover the core meaning but omit relevant sociocultural nuances, such as a term's association with specific African-American or Latinx queer communities, its reclaimed status, or its potentially harmful use outside in-group contexts.

Finally, the analysis highlights some limitations of the automatic evaluation. Correct definitions may receive lower similarity scores when models include another valid sense, explain the sentence context, or provide reasoning before reaching the intended meaning. For example, a model may correctly identify the slang meaning of a polysemous term while also supplying its literal meaning. The manual labels therefore distinguish genuine failures of slang understanding from verbosity, valid handling of polysemy, and correct definitions that differ in scope or formulation from the available references. 
Appendix~\ref{app:manual-evaluation} presents selected examples from the qualitative analysis, while the complete manual annotations are available in the paper's repository.

{\section{Conclusion}

This work presents an initial evaluation of LLMs on queer slang in user-generated content. We introduced \textsc{Slang-Q}, a manually curated dataset of 1,024 English sentences paired with queer slang terms and reference definitions.
The current stage of the research primarily focuses on data creation, dataset analysis, and a first round of experiments on a small set of models. 
Our first exploratory experiments show that models fall below the human upper bound, that withholding domain framing and
sentential context consistently hurts performance, and that both  slang-informed prompting and contextual grounding help models converge on the intended queer-specific meaning. The manual evaluation supports this finding: context eliminates fully incorrect outputs in the analyzed sample, although some definitions still omit culturally specific or pragmatic nuances. 

This work has several limitations. First, the dataset is limited to English user-generated content, while queer slang varies considerably across languages, cultures, and platforms. Our evaluation focuses on definition generation; future work could broaden the range of tasks to include slang-aware dialogue, summarization, translation, and other forms of content understanding. A larger-scale manual assessment could
also examine variation across term categories and historical periods, to examine whether models perform differently on established versus recently emerged slang. Finally, we note that the adopted definitions and categorizations represent a snapshot as of 2026. As grassroots, community-maintained projects, such lexicons are inherently dynamic: terms, meanings, and category assignments may evolve over time and be subject to ongoing debate within the communities that use them.

\section*{Acknowledgements}
The work by Beatrice Savoldi is supported by the European Union’s Horizon Europe programme under grant agreement No. 101213369 (DVPS).

\bibliography{references}

\begin{thebibliography}{30}
\expandafter\ifx\csname natexlab\endcsname\relax\def\natexlab#1{#1}\fi
\providecommand{\url}[1]{\texttt{#1}}
\providecommand{\href}[2]{#2}
\providecommand{\path}[1]{#1}
\providecommand{\DOIprefix}{doi:}
\providecommand{\ArXivprefix}{arXiv:}
\providecommand{\URLprefix}{URL: }
\providecommand{\Pubmedprefix}{pmid:}
\providecommand{\doi}[1]{\href{http://dx.doi.org/#1}{\path{#1}}}
\providecommand{\Pubmed}[1]{\href{pmid:#1}{\path{#1}}}
\providecommand{\bibinfo}[2]{#2}
\ifx\xfnm\relax \def\xfnm[#1]{\unskip,\space#1}\fi
\bibitem[{Chen and Shu(2024)}]{chen2024combating}
\bibinfo{author}{C.~Chen}, \bibinfo{author}{K.~Shu},
\newblock \bibinfo{title}{Combating misinformation in the age of llms: Opportunities and challenges},
\newblock \bibinfo{journal}{AI magazine} \bibinfo{volume}{45} (\bibinfo{year}{2024}) \bibinfo{pages}{354--368}.
\bibitem[{Chatterji et~al.(2025)Chatterji, Cunningham, Deming, Hitzig, Ong, Shan, and Wadman}]{chatterji2025people}
\bibinfo{author}{A.~Chatterji}, \bibinfo{author}{T.~Cunningham}, \bibinfo{author}{D.~J. Deming}, \bibinfo{author}{Z.~Hitzig}, \bibinfo{author}{C.~Ong}, \bibinfo{author}{C.~Y. Shan}, \bibinfo{author}{K.~Wadman}, \bibinfo{title}{How people use chatgpt}, \bibinfo{type}{Technical Report}, National Bureau of Economic Research, \bibinfo{year}{2025}.
\bibitem[{Savoldi et~al.(2026)Savoldi, Attanasio, Gorodetskaya, Manerba, Bassignana, Casola, Negri, Caselli, Bentivogli, Ramponi, Muti, Balbo, and Nozza}]{savoldi2026generativeaipracticesliteracy}
\bibinfo{author}{B.~Savoldi}, \bibinfo{author}{G.~Attanasio}, \bibinfo{author}{O.~Gorodetskaya}, \bibinfo{author}{M.~M. Manerba}, \bibinfo{author}{E.~Bassignana}, \bibinfo{author}{S.~Casola}, \bibinfo{author}{M.~Negri}, \bibinfo{author}{T.~Caselli}, \bibinfo{author}{L.~Bentivogli}, \bibinfo{author}{A.~Ramponi}, \bibinfo{author}{A.~Muti}, \bibinfo{author}{N.~Balbo}, \bibinfo{author}{D.~Nozza}, \bibinfo{title}{Generative ai practices, literacy, and divides: An empirical analysis in the italian context}, \bibinfo{year}{2026}. \URLprefix \url{https://arxiv.org/abs/2512.03671}. \href{http://arxiv.org/abs/2512.03671}{{\tt arXiv:2512.03671}}.
\bibitem[{Basoah et~al.(2025)Basoah, Chechelnitsky, Long, Reinecke, Zerva, Zhou, D{\'\i}az, and Sap}]{basoah2025not}
\bibinfo{author}{J.~Basoah}, \bibinfo{author}{D.~Chechelnitsky}, \bibinfo{author}{T.~Long}, \bibinfo{author}{K.~Reinecke}, \bibinfo{author}{C.~Zerva}, \bibinfo{author}{K.~Zhou}, \bibinfo{author}{M.~D{\'\i}az}, \bibinfo{author}{M.~Sap},
\newblock \bibinfo{title}{Not like us, hunty: Measuring perceptions and behavioral effects of minoritized anthropomorphic cues in llms},
\newblock in: \bibinfo{booktitle}{Proceedings of the 2025 ACM Conference on Fairness, Accountability, and Transparency}, \bibinfo{year}{2025}, pp. \bibinfo{pages}{710--745}.
\bibitem[{Mei et~al.(2024)Mei, Liu, Wang, Bi, and Cheng}]{mei-etal-2024-slang}
\bibinfo{author}{L.~Mei}, \bibinfo{author}{S.~Liu}, \bibinfo{author}{Y.~Wang}, \bibinfo{author}{B.~Bi}, \bibinfo{author}{X.~Cheng},
\newblock \bibinfo{title}{{SLANG}: New concept comprehension of large language models},
\newblock in: \bibinfo{editor}{Y.~Al-Onaizan}, \bibinfo{editor}{M.~Bansal}, \bibinfo{editor}{Y.-N. Chen} (Eds.), \bibinfo{booktitle}{Proceedings of the 2024 Conference on Empirical Methods in Natural Language Processing}, \bibinfo{publisher}{Association for Computational Linguistics}, \bibinfo{address}{Miami, Florida, USA}, \bibinfo{year}{2024}, pp. \bibinfo{pages}{12558--12575}. \URLprefix \url{https://aclanthology.org/2024.emnlp-main.698/}. \DOIprefix\doi{10.18653/v1/2024.emnlp-main.698}.
\bibitem[{Eble(1996)}]{eble1996slang}
\bibinfo{author}{C.~C. Eble}, \bibinfo{title}{Slang \& sociability: In-group language among college students}, \bibinfo{publisher}{Univ of North Carolina Press}, \bibinfo{year}{1996}.
\bibitem[{Kulick(2000)}]{kulick2000gay}
\bibinfo{author}{D.~Kulick},
\newblock \bibinfo{title}{Gay and lesbian language},
\newblock \bibinfo{journal}{Annual review of anthropology} \bibinfo{volume}{29} (\bibinfo{year}{2000}) \bibinfo{pages}{243--285}.
\bibitem[{Motschenbacher(2011)}]{motschenbacher2011taking}
\bibinfo{author}{H.~Motschenbacher},
\newblock \bibinfo{title}{Taking queer linguistics further: Sociolinguistics and critical heteronormativity research.},
\newblock \bibinfo{journal}{International journal of the sociology of language} \bibinfo{volume}{2011} (\bibinfo{year}{2011}).
\bibitem[{Cifor and Rawson(2023)}]{cifor2023mediating}
\bibinfo{author}{M.~Cifor}, \bibinfo{author}{K.~Rawson},
\newblock \bibinfo{title}{Mediating queer and trans pasts: The homosaurus as queer information activism},
\newblock \bibinfo{journal}{Information, Communication \& Society} \bibinfo{volume}{26} (\bibinfo{year}{2023}) \bibinfo{pages}{2168--2185}.
\bibitem[{Weber et~al.(2026)Weber, Wang, Gupta, Subramonian, Ulmer, Tanwar, Aich, Devinney, Hobbs, Mickel et~al.}]{weber2026queer}
\bibinfo{author}{S.~Weber}, \bibinfo{author}{A.~Wang}, \bibinfo{author}{A.~Gupta}, \bibinfo{author}{A.~Subramonian}, \bibinfo{author}{D.~Ulmer}, \bibinfo{author}{E.~Tanwar}, \bibinfo{author}{G.~Aich}, \bibinfo{author}{H.~Devinney}, \bibinfo{author}{J.~Hobbs}, \bibinfo{author}{J.~Mickel}, et~al.,
\newblock \bibinfo{title}{Queer nlp: A critical survey on literature gaps, biases and trends},
\newblock \bibinfo{journal}{arXiv preprint arXiv:2602.16151}  (\bibinfo{year}{2026}).
\bibitem[{Dorn et~al.(2024)Dorn, Kezar, Morstatter, and Lerman}]{dorn2024harmful}
\bibinfo{author}{R.~Dorn}, \bibinfo{author}{L.~Kezar}, \bibinfo{author}{F.~Morstatter}, \bibinfo{author}{K.~Lerman},
\newblock \bibinfo{title}{Harmful speech detection by language models exhibits gender-queer dialect bias},
\newblock in: \bibinfo{booktitle}{Proceedings of the 4th ACM Conference on Equity and Access in Algorithms, Mechanisms, and Optimization}, \bibinfo{year}{2024}, pp. \bibinfo{pages}{1--12}.
\bibitem[{V{\'a}squez et~al.(2022)V{\'a}squez, Bel-Enguix, Andersen, and Ojeda-Trueba}]{vasquez2022heterocorpus}
\bibinfo{author}{J.~V{\'a}squez}, \bibinfo{author}{G.~Bel-Enguix}, \bibinfo{author}{S.~T. Andersen}, \bibinfo{author}{S.-L. Ojeda-Trueba},
\newblock \bibinfo{title}{Heterocorpus: A corpus for heteronormative language detection},
\newblock in: \bibinfo{booktitle}{Proceedings of the 4th Workshop on Gender Bias in Natural Language Processing (GeBNLP)}, \bibinfo{year}{2022}, pp. \bibinfo{pages}{225--234}.
\bibitem[{Tint(2025)}]{tint-2025-guardrails}
\bibinfo{author}{J.~Tint},
\newblock \bibinfo{title}{Guardrails, not guidance: Understanding responses to {LGBTQ}+ language in large language models},
\newblock in: \bibinfo{editor}{A.~Pranav}, \bibinfo{editor}{A.~Valentine}, \bibinfo{editor}{S.~Bhatt}, \bibinfo{editor}{Y.~Long}, \bibinfo{editor}{A.~Subramonian}, \bibinfo{editor}{A.~Bertsch}, \bibinfo{editor}{A.~Lauscher}, \bibinfo{editor}{A.~Gupta} (Eds.), \bibinfo{booktitle}{Proceedings of the Queer in AI Workshop}, \bibinfo{publisher}{Association for Computational Linguistics}, \bibinfo{address}{Hybrid format (in-person and virtual)}, \bibinfo{year}{2025}, pp. \bibinfo{pages}{6--16}. \URLprefix \url{https://aclanthology.org/2025.queerinai-main.2/}. \DOIprefix\doi{10.18653/v1/2025.queerinai-main.2}.
\bibitem[{Weber et~al.(2026)Weber, Wang, Gupta, Subramonian, Ulmer, Tanwar, Aich, Devinney, Hobbs, Mickel, Tint, Sosto, Groshan, Astarita, Gautam, Blaschke, Agnew, Lee, and Long}]{weber2026queernlpcriticalsurvey}
\bibinfo{author}{S.~Weber}, \bibinfo{author}{A.~Wang}, \bibinfo{author}{A.~Gupta}, \bibinfo{author}{A.~Subramonian}, \bibinfo{author}{D.~Ulmer}, \bibinfo{author}{E.~Tanwar}, \bibinfo{author}{G.~Aich}, \bibinfo{author}{H.~Devinney}, \bibinfo{author}{J.~Hobbs}, \bibinfo{author}{J.~Mickel}, \bibinfo{author}{J.~Tint}, \bibinfo{author}{M.~Sosto}, \bibinfo{author}{R.~Groshan}, \bibinfo{author}{S.~Astarita}, \bibinfo{author}{V.~Gautam}, \bibinfo{author}{V.~Blaschke}, \bibinfo{author}{W.~Agnew}, \bibinfo{author}{W.~Y. Lee}, \bibinfo{author}{Y.~Long}, \bibinfo{title}{Queer nlp: A critical survey on literature gaps, biases and trends}, \bibinfo{year}{2026}. \URLprefix \url{https://arxiv.org/abs/2602.16151}. \href{http://arxiv.org/abs/2602.16151}{{\tt arXiv:2602.16151}}.
\bibitem[{Sosto et~al.(2026)Sosto, Pandiani, and Hollink}]{sosto-etal-2026-queergen}
\bibinfo{author}{M.~Sosto}, \bibinfo{author}{D.~S.~M. Pandiani}, \bibinfo{author}{L.~Hollink},
\newblock \bibinfo{title}{{Q}ueer{G}en: How {LLM}s reflect societal norms on gender and sexuality in sentence completion task},
\newblock in: \bibinfo{editor}{V.~Demberg}, \bibinfo{editor}{K.~Inui}, \bibinfo{editor}{L.~Marquez} (Eds.), \bibinfo{booktitle}{Findings of the {A}ssociation for {C}omputational {L}inguistics: {EACL} 2026}, \bibinfo{publisher}{Association for Computational Linguistics}, \bibinfo{address}{Rabat, Morocco}, \bibinfo{year}{2026}, pp. \bibinfo{pages}{4305--4326}. \URLprefix \url{https://aclanthology.org/2026.findings-eacl.225/}. \DOIprefix\doi{10.18653/v1/2026.findings-eacl.225}.
\bibitem[{{Library.LGBT}(nd)}]{librarylgbt}
\bibinfo{author}{{Library.LGBT}}, \bibinfo{title}{Lexicon library.lgbt}, \bibinfo{year}{n.d.} \URLprefix \url{https://lexicon.library.lgbt/}.
\bibitem[{Veloso et~al.(2025)Veloso, Hirlimann, Wicke, and Schütze}]{veloso2025slayingqueerlanguageprocessing}
\bibinfo{author}{L.~Veloso}, \bibinfo{author}{L.~Hirlimann}, \bibinfo{author}{P.~Wicke}, \bibinfo{author}{H.~Schütze}, \bibinfo{title}{Slaying: Towards queer language processing}, \bibinfo{year}{2025}. \URLprefix \url{https://arxiv.org/abs/2509.17449}. \href{http://arxiv.org/abs/2509.17449}{{\tt arXiv:2509.17449}}.
\bibitem[{Piergentili et~al.(2024)Piergentili, Savoldi, Negri, and Bentivogli}]{piergentili-etal-2024-enhancing}
\bibinfo{author}{A.~Piergentili}, \bibinfo{author}{B.~Savoldi}, \bibinfo{author}{M.~Negri}, \bibinfo{author}{L.~Bentivogli},
\newblock \bibinfo{title}{Enhancing gender-inclusive machine translation with neomorphemes and large language models},
\newblock in: \bibinfo{editor}{C.~Scarton}, \bibinfo{editor}{C.~Prescott}, \bibinfo{editor}{C.~Bayliss}, \bibinfo{editor}{C.~Oakley}, \bibinfo{editor}{J.~Wright}, \bibinfo{editor}{S.~Wrigley}, \bibinfo{editor}{X.~Song}, \bibinfo{editor}{E.~Gow-Smith}, \bibinfo{editor}{R.~Bawden}, \bibinfo{editor}{V.~M. S{\'a}nchez-Cartagena}, \bibinfo{editor}{P.~Cadwell}, \bibinfo{editor}{E.~Lapshinova-Koltunski}, \bibinfo{editor}{V.~Cabarr{\~a}o}, \bibinfo{editor}{K.~Chatzitheodorou}, \bibinfo{editor}{M.~Nurminen}, \bibinfo{editor}{D.~Kanojia}, \bibinfo{editor}{H.~Moniz} (Eds.), \bibinfo{booktitle}{Proceedings of the 25th Annual Conference of the European Association for Machine Translation (Volume 1)}, \bibinfo{publisher}{European Association for Machine Translation (EAMT)}, \bibinfo{address}{Sheffield, UK}, \bibinfo{year}{2024}, pp. \bibinfo{pages}{300--314}. \URLprefix \url{https://aclanthology.org/2024.eamt-1.25/}.
\bibitem[{Ungless et~al.(2025)Ungless, Dev, Bennett, Gulotta, Bastings, and Denton}]{ungless-etal-2025-amplifying}
\bibinfo{author}{E.~L. Ungless}, \bibinfo{author}{S.~Dev}, \bibinfo{author}{C.~L. Bennett}, \bibinfo{author}{R.~Gulotta}, \bibinfo{author}{J.~Bastings}, \bibinfo{author}{R.~Denton},
\newblock \bibinfo{title}{Amplifying trans and nonbinary voices: A community-centred harm taxonomy for {LLM}s},
\newblock in: \bibinfo{editor}{W.~Che}, \bibinfo{editor}{J.~Nabende}, \bibinfo{editor}{E.~Shutova}, \bibinfo{editor}{M.~T. Pilehvar} (Eds.), \bibinfo{booktitle}{Proceedings of the 63rd Annual Meeting of the Association for Computational Linguistics (Volume 1: Long Papers)}, \bibinfo{publisher}{Association for Computational Linguistics}, \bibinfo{address}{Vienna, Austria}, \bibinfo{year}{2025}, pp. \bibinfo{pages}{20503--20535}. \URLprefix \url{https://aclanthology.org/2025.acl-long.1001/}. \DOIprefix\doi{10.18653/v1/2025.acl-long.1001}.
\bibitem[{Locatelli et~al.(2023)Locatelli, Damo, and Nozza}]{locatelli-etal-2023-cross}
\bibinfo{author}{D.~Locatelli}, \bibinfo{author}{G.~Damo}, \bibinfo{author}{D.~Nozza},
\newblock \bibinfo{title}{A cross-lingual study of homotransphobia on {T}witter},
\newblock in: \bibinfo{editor}{S.~Dev}, \bibinfo{editor}{V.~Prabhakaran}, \bibinfo{editor}{D.~I. Adelani}, \bibinfo{editor}{D.~Hovy}, \bibinfo{editor}{L.~Benotti} (Eds.), \bibinfo{booktitle}{Proceedings of the First Workshop on Cross-Cultural Considerations in NLP (C3NLP)}, \bibinfo{publisher}{Association for Computational Linguistics}, \bibinfo{address}{Dubrovnik, Croatia}, \bibinfo{year}{2023}, pp. \bibinfo{pages}{16--24}. \URLprefix \url{https://aclanthology.org/2023.c3nlp-1.3/}. \DOIprefix\doi{10.18653/v1/2023.c3nlp-1.3}.
\bibitem[{Pofcher et~al.(2025)Pofcher, Homan, Sell, and KhudaBukhsh}]{pofcher-etal-2025-hope}
\bibinfo{author}{J.~Pofcher}, \bibinfo{author}{C.~M. Homan}, \bibinfo{author}{R.~Sell}, \bibinfo{author}{A.~R. KhudaBukhsh},
\newblock \bibinfo{title}{Hope vs. hate: Understanding user interactions with {LGBTQ}+ news content in mainstream {US} news media through the lens of hope speech},
\newblock in: \bibinfo{editor}{C.~Christodoulopoulos}, \bibinfo{editor}{T.~Chakraborty}, \bibinfo{editor}{C.~Rose}, \bibinfo{editor}{V.~Peng} (Eds.), \bibinfo{booktitle}{Proceedings of the 2025 Conference on Empirical Methods in Natural Language Processing}, \bibinfo{publisher}{Association for Computational Linguistics}, \bibinfo{address}{Suzhou, China}, \bibinfo{year}{2025}, pp. \bibinfo{pages}{19862--19888}. \URLprefix \url{https://aclanthology.org/2025.emnlp-main.1005/}. \DOIprefix\doi{10.18653/v1/2025.emnlp-main.1005}.
\bibitem[{Ghosal et~al.(2025)Ghosal, Gupta, and Srikumar}]{ghosal2025unequalvoicesllmsconstruct}
\bibinfo{author}{A.~Ghosal}, \bibinfo{author}{A.~Gupta}, \bibinfo{author}{V.~Srikumar}, \bibinfo{title}{Unequal voices: How llms construct constrained queer narratives}, \bibinfo{year}{2025}. \URLprefix \url{https://arxiv.org/abs/2507.15585}. \href{http://arxiv.org/abs/2507.15585}{{\tt arXiv:2507.15585}}.
\bibitem[{Ramesh et~al.(2022)Ramesh, Kumar, and Khudabukhsh}]{ramesh-etal-2022-revisiting}
\bibinfo{author}{K.~Ramesh}, \bibinfo{author}{S.~Kumar}, \bibinfo{author}{A.~Khudabukhsh},
\newblock \bibinfo{title}{Revisiting queer minorities in lexicons},
\newblock in: \bibinfo{editor}{K.~Narang}, \bibinfo{editor}{A.~Mostafazadeh~Davani}, \bibinfo{editor}{L.~Mathias}, \bibinfo{editor}{B.~Vidgen}, \bibinfo{editor}{Z.~Talat} (Eds.), \bibinfo{booktitle}{Proceedings of the Sixth Workshop on Online Abuse and Harms (WOAH)}, \bibinfo{publisher}{Association for Computational Linguistics}, \bibinfo{address}{Seattle, Washington (Hybrid)}, \bibinfo{year}{2022}, pp. \bibinfo{pages}{245--251}. \URLprefix \url{https://aclanthology.org/2022.woah-1.23/}. \DOIprefix\doi{10.18653/v1/2022.woah-1.23}.
\bibitem[{Subramonian et~al.(2025)Subramonian, Gautam, Seshadri, Klakow, Chang, and Sun}]{subramonian2025agreedisagreemetaevaluationllm}
\bibinfo{author}{A.~Subramonian}, \bibinfo{author}{V.~Gautam}, \bibinfo{author}{P.~Seshadri}, \bibinfo{author}{D.~Klakow}, \bibinfo{author}{K.-W. Chang}, \bibinfo{author}{Y.~Sun}, \bibinfo{title}{Agree to disagree? a meta-evaluation of llm misgendering}, \bibinfo{year}{2025}. \URLprefix \url{https://arxiv.org/abs/2504.17075}. \href{http://arxiv.org/abs/2504.17075}{{\tt arXiv:2504.17075}}.
\bibitem[{Team(2025)}]{qwen3technicalreport}
\bibinfo{author}{Q.~Team}, \bibinfo{title}{Qwen3 technical report}, \bibinfo{year}{2025}. \URLprefix \url{https://arxiv.org/abs/2505.09388}. \href{http://arxiv.org/abs/2505.09388}{{\tt arXiv:2505.09388}}.
\bibitem[{by~arXiv(2026)}]{arxiv2026llama4herdarchitecture}
\bibinfo{author}{R.~by~arXiv}, \bibinfo{title}{The llama 4 herd: Architecture, training, evaluation, and deployment notes}, \bibinfo{year}{2026}. \URLprefix \url{https://arxiv.org/abs/2601.11659}. \href{http://arxiv.org/abs/2601.11659}{{\tt arXiv:2601.11659}}.
\bibitem[{Grattafiori et~al.(2024)Grattafiori, Dubey, Jauhri, Pandey, Kadian, Al-Dahle, Letman, Mathur, Schelten, Vaughan, Yang, Fan, Goyal, Hartshorn, Yang, Mitra, Sravankumar, Korenev, Hinsvark, Rao, Zhang, Rodriguez, Gregerson, Spataru, Roziere, Biron, Tang, Chern, Caucheteux, Nayak, Bi, Marra, McConnell, Keller, Touret, Wu, Wong, Ferrer, Nikolaidis, Allonsius, Song, Pintz, Livshits, Wyatt, Esiobu, Choudhary, Mahajan, Garcia-Olano, Perino, Hupkes, Lakomkin, AlBadawy, Lobanova, Dinan, Smith, Radenovic, Guzmán, Zhang, Synnaeve, Lee, Anderson, Thattai, Nail, Mialon, Pang, Cucurell, Nguyen, Korevaar, Xu, Touvron, Zarov, Ibarra, Kloumann, Misra, Evtimov, Zhang, Copet, Lee, Geffert, Vranes, Park, Mahadeokar, Shah, van~der Linde, Billock, Hong, Lee, Fu, Chi, Huang, Liu, Wang, Yu, Bitton, Spisak, Park, Rocca, Johnstun, Saxe, Jia, Alwala, Prasad, Upasani, Plawiak, Li, Heafield, Stone, El-Arini, Iyer, Malik, Chiu, Bhalla, Lakhotia, Rantala-Yeary, van~der Maaten, Chen, Tan, Jenkins, Martin, Madaan, Malo, Blecher,
  Landzaat, de~Oliveira, Muzzi, Pasupuleti, Singh, Paluri, Kardas, Tsimpoukelli, Oldham, Rita, Pavlova, Kambadur, Lewis, Si, Singh, Hassan, Goyal, Torabi, Bashlykov, Bogoychev, Chatterji, Zhang, Duchenne, Çelebi, Alrassy, Zhang, Li, Vasic, Weng, Bhargava, Dubal, Krishnan, Koura, Xu, He, Dong, Srinivasan, Ganapathy, Calderer, Cabral, Stojnic, Raileanu, Maheswari, Girdhar, Patel, Sauvestre, Polidoro, Sumbaly, Taylor, Silva, Hou, Wang, Hosseini, Chennabasappa, Singh, Bell, Kim, Edunov, Nie, Narang, Raparthy, Shen, Wan, Bhosale, Zhang, Vandenhende, Batra, Whitman, Sootla, Collot, Gururangan, Borodinsky, Herman, Fowler, Sheasha, Georgiou, Scialom, Speckbacher, Mihaylov, Xiao, Karn, Goswami, Gupta, Ramanathan, Kerkez, Gonguet, Do, Vogeti, Albiero, Petrovic, Chu, Xiong, Fu, Meers, Martinet, Wang, Wang, Tan, Xia, Xie, Jia, Wang, Goldschlag, Gaur, Babaei, Wen, Song, Zhang, Li, Mao, Coudert, Yan, Chen, Papakipos, Singh, Srivastava, Jain, Kelsey, Shajnfeld, Gangidi, Victoria, Goldstand, Menon, Sharma, Boesenberg,
  Baevski, Feinstein, Kallet, Sangani, Teo, Yunus, Lupu, Alvarado, Caples, Gu, Ho, Poulton, Ryan, Ramchandani, Dong, Franco, Goyal, Saraf, Chowdhury, Gabriel, Bharambe, Eisenman, Yazdan, James, Maurer, Leonhardi, Huang, Loyd, Paola, Paranjape, Liu, Wu, Ni, Hancock, Wasti, Spence, Stojkovic, Gamido, Montalvo, Parker, Burton, Mejia, Liu, Wang, Kim, Zhou, Hu, Chu, Cai, Tindal, Feichtenhofer, Gao, Civin, Beaty, Kreymer, Li, Adkins, Xu, Testuggine, David, Parikh, Liskovich, Foss, Wang, Le, Holland, Dowling, Jamil, Montgomery, Presani, Hahn, Wood, Le, Brinkman, Arcaute, Dunbar, Smothers, Sun, Kreuk, Tian, Kokkinos, Ozgenel, Caggioni, Kanayet, Seide, Florez, Schwarz, Badeer, Swee, Halpern, Herman, Sizov, Guangyi, Zhang, Lakshminarayanan, Inan, Shojanazeri, Zou, Wang, Zha, Habeeb, Rudolph, Suk, Aspegren, Goldman, Zhan, Damlaj, Molybog, Tufanov, Leontiadis, Veliche, Gat, Weissman, Geboski, Kohli, Lam, Asher, Gaya, Marcus, Tang, Chan, Zhen, Reizenstein, Teboul, Zhong, Jin, Yang, Cummings, Carvill, Shepard, McPhie,
  Torres, Ginsburg, Wang, Wu, U, Saxena, Khandelwal, Zand, Matosich, Veeraraghavan, Michelena, Li, Jagadeesh, Huang, Chawla, Huang, Chen, Garg, A, Silva, Bell, Zhang, Guo, Yu, Moshkovich, Wehrstedt, Khabsa, Avalani, Bhatt, Mankus, Hasson, Lennie, Reso, Groshev, Naumov, Lathi, Keneally, Liu, Seltzer, Valko, Restrepo, Patel, Vyatskov, Samvelyan, Clark, Macey, Wang, Hermoso, Metanat, Rastegari, Bansal, Santhanam, Parks, White, Bawa, Singhal, Egebo, Usunier, Mehta, Laptev, Dong, Cheng, Chernoguz, Hart, Salpekar, Kalinli, Kent, Parekh, Saab, Balaji, Rittner, Bontrager, Roux, Dollar, Zvyagina, Ratanchandani, Yuvraj, Liang, Alao, Rodriguez, Ayub, Murthy, Nayani, Mitra, Parthasarathy, Li, Hogan, Battey, Wang, Howes, Rinott, Mehta, Siby, Bondu, Datta, Chugh, Hunt, Dhillon, Sidorov, Pan, Mahajan, Verma, Yamamoto, Ramaswamy, Lindsay, Lindsay, Feng, Lin, Zha, Patil, Shankar, Zhang, Zhang, Wang, Agarwal, Sajuyigbe, Chintala, Max, Chen, Kehoe, Satterfield, Govindaprasad, Gupta, Deng, Cho, Virk, Subramanian, Choudhury,
  Goldman, Remez, Glaser, Best, Koehler, Robinson, Li, Zhang, Matthews, Chou, Shaked, Vontimitta, Ajayi, Montanez, Mohan, Kumar, Mangla, Ionescu, Poenaru, Mihailescu, Ivanov, Li, Wang, Jiang, Bouaziz, Constable, Tang, Wu, Wang, Wu, Gao, Kleinman, Chen, Hu, Jia, Qi, Li, Zhang, Zhang, Adi, Nam, Yu, Wang, Zhao, Hao, Qian, Li, He, Rait, DeVito, Rosnbrick, Wen, Yang, Zhao, and Ma}]{grattafiori2024llama3herdmodels}
\bibinfo{author}{A.~Grattafiori}, \bibinfo{author}{A.~Dubey}, \bibinfo{author}{A.~Jauhri}, \bibinfo{author}{A.~Pandey}, \bibinfo{author}{A.~Kadian}, \bibinfo{author}{A.~Al-Dahle}, \bibinfo{author}{A.~Letman}, \bibinfo{author}{A.~Mathur}, \bibinfo{author}{A.~Schelten}, \bibinfo{author}{A.~Vaughan}, \bibinfo{author}{A.~Yang}, \bibinfo{author}{A.~Fan}, \bibinfo{author}{A.~Goyal}, \bibinfo{author}{A.~Hartshorn}, \bibinfo{author}{A.~Yang}, \bibinfo{author}{A.~Mitra}, \bibinfo{author}{A.~Sravankumar}, \bibinfo{author}{A.~Korenev}, \bibinfo{author}{A.~Hinsvark}, \bibinfo{author}{A.~Rao}, \bibinfo{author}{A.~Zhang}, \bibinfo{author}{A.~Rodriguez}, \bibinfo{author}{A.~Gregerson}, \bibinfo{author}{A.~Spataru}, \bibinfo{author}{B.~Roziere}, \bibinfo{author}{B.~Biron}, \bibinfo{author}{B.~Tang}, \bibinfo{author}{B.~Chern}, \bibinfo{author}{C.~Caucheteux}, \bibinfo{author}{C.~Nayak}, \bibinfo{author}{C.~Bi}, \bibinfo{author}{C.~Marra}, \bibinfo{author}{C.~McConnell}, \bibinfo{author}{C.~Keller},
  \bibinfo{author}{C.~Touret}, \bibinfo{author}{C.~Wu}, \bibinfo{author}{C.~Wong}, \bibinfo{author}{C.~C. Ferrer}, \bibinfo{author}{C.~Nikolaidis}, \bibinfo{author}{D.~Allonsius}, \bibinfo{author}{D.~Song}, \bibinfo{author}{D.~Pintz}, \bibinfo{author}{D.~Livshits}, \bibinfo{author}{D.~Wyatt}, \bibinfo{author}{D.~Esiobu}, \bibinfo{author}{D.~Choudhary}, \bibinfo{author}{D.~Mahajan}, \bibinfo{author}{D.~Garcia-Olano}, \bibinfo{author}{D.~Perino}, \bibinfo{author}{D.~Hupkes}, \bibinfo{author}{E.~Lakomkin}, \bibinfo{author}{E.~AlBadawy}, \bibinfo{author}{E.~Lobanova}, \bibinfo{author}{E.~Dinan}, \bibinfo{author}{E.~M. Smith}, \bibinfo{author}{F.~Radenovic}, \bibinfo{author}{F.~Guzmán}, \bibinfo{author}{F.~Zhang}, \bibinfo{author}{G.~Synnaeve}, \bibinfo{author}{G.~Lee}, \bibinfo{author}{G.~L. Anderson}, \bibinfo{author}{G.~Thattai}, \bibinfo{author}{G.~Nail}, \bibinfo{author}{G.~Mialon}, \bibinfo{author}{G.~Pang}, \bibinfo{author}{G.~Cucurell}, \bibinfo{author}{H.~Nguyen}, \bibinfo{author}{H.~Korevaar},
  \bibinfo{author}{H.~Xu}, \bibinfo{author}{H.~Touvron}, \bibinfo{author}{I.~Zarov}, \bibinfo{author}{I.~A. Ibarra}, \bibinfo{author}{I.~Kloumann}, \bibinfo{author}{I.~Misra}, \bibinfo{author}{I.~Evtimov}, \bibinfo{author}{J.~Zhang}, \bibinfo{author}{J.~Copet}, \bibinfo{author}{J.~Lee}, \bibinfo{author}{J.~Geffert}, \bibinfo{author}{J.~Vranes}, \bibinfo{author}{J.~Park}, \bibinfo{author}{J.~Mahadeokar}, \bibinfo{author}{J.~Shah}, \bibinfo{author}{J.~van~der Linde}, \bibinfo{author}{J.~Billock}, \bibinfo{author}{J.~Hong}, \bibinfo{author}{J.~Lee}, \bibinfo{author}{J.~Fu}, \bibinfo{author}{J.~Chi}, \bibinfo{author}{J.~Huang}, \bibinfo{author}{J.~Liu}, \bibinfo{author}{J.~Wang}, \bibinfo{author}{J.~Yu}, \bibinfo{author}{J.~Bitton}, \bibinfo{author}{J.~Spisak}, \bibinfo{author}{J.~Park}, \bibinfo{author}{J.~Rocca}, \bibinfo{author}{J.~Johnstun}, \bibinfo{author}{J.~Saxe}, \bibinfo{author}{J.~Jia}, \bibinfo{author}{K.~V. Alwala}, \bibinfo{author}{K.~Prasad}, \bibinfo{author}{K.~Upasani},
  \bibinfo{author}{K.~Plawiak}, \bibinfo{author}{K.~Li}, \bibinfo{author}{K.~Heafield}, \bibinfo{author}{K.~Stone}, \bibinfo{author}{K.~El-Arini}, \bibinfo{author}{K.~Iyer}, \bibinfo{author}{K.~Malik}, \bibinfo{author}{K.~Chiu}, \bibinfo{author}{K.~Bhalla}, \bibinfo{author}{K.~Lakhotia}, \bibinfo{author}{L.~Rantala-Yeary}, \bibinfo{author}{L.~van~der Maaten}, \bibinfo{author}{L.~Chen}, \bibinfo{author}{L.~Tan}, \bibinfo{author}{L.~Jenkins}, \bibinfo{author}{L.~Martin}, \bibinfo{author}{L.~Madaan}, \bibinfo{author}{L.~Malo}, \bibinfo{author}{L.~Blecher}, \bibinfo{author}{L.~Landzaat}, \bibinfo{author}{L.~de~Oliveira}, \bibinfo{author}{M.~Muzzi}, \bibinfo{author}{M.~Pasupuleti}, \bibinfo{author}{M.~Singh}, \bibinfo{author}{M.~Paluri}, \bibinfo{author}{M.~Kardas}, \bibinfo{author}{M.~Tsimpoukelli}, \bibinfo{author}{M.~Oldham}, \bibinfo{author}{M.~Rita}, \bibinfo{author}{M.~Pavlova}, \bibinfo{author}{M.~Kambadur}, \bibinfo{author}{M.~Lewis}, \bibinfo{author}{M.~Si}, \bibinfo{author}{M.~K. Singh},
  \bibinfo{author}{M.~Hassan}, \bibinfo{author}{N.~Goyal}, \bibinfo{author}{N.~Torabi}, \bibinfo{author}{N.~Bashlykov}, \bibinfo{author}{N.~Bogoychev}, \bibinfo{author}{N.~Chatterji}, \bibinfo{author}{N.~Zhang}, \bibinfo{author}{O.~Duchenne}, \bibinfo{author}{O.~Çelebi}, \bibinfo{author}{P.~Alrassy}, \bibinfo{author}{P.~Zhang}, \bibinfo{author}{P.~Li}, \bibinfo{author}{P.~Vasic}, \bibinfo{author}{P.~Weng}, \bibinfo{author}{P.~Bhargava}, \bibinfo{author}{P.~Dubal}, \bibinfo{author}{P.~Krishnan}, \bibinfo{author}{P.~S. Koura}, \bibinfo{author}{P.~Xu}, \bibinfo{author}{Q.~He}, \bibinfo{author}{Q.~Dong}, \bibinfo{author}{R.~Srinivasan}, \bibinfo{author}{R.~Ganapathy}, \bibinfo{author}{R.~Calderer}, \bibinfo{author}{R.~S. Cabral}, \bibinfo{author}{R.~Stojnic}, \bibinfo{author}{R.~Raileanu}, \bibinfo{author}{R.~Maheswari}, \bibinfo{author}{R.~Girdhar}, \bibinfo{author}{R.~Patel}, \bibinfo{author}{R.~Sauvestre}, \bibinfo{author}{R.~Polidoro}, \bibinfo{author}{R.~Sumbaly}, \bibinfo{author}{R.~Taylor},
  \bibinfo{author}{R.~Silva}, \bibinfo{author}{R.~Hou}, \bibinfo{author}{R.~Wang}, \bibinfo{author}{S.~Hosseini}, \bibinfo{author}{S.~Chennabasappa}, \bibinfo{author}{S.~Singh}, \bibinfo{author}{S.~Bell}, \bibinfo{author}{S.~S. Kim}, \bibinfo{author}{S.~Edunov}, \bibinfo{author}{S.~Nie}, \bibinfo{author}{S.~Narang}, \bibinfo{author}{S.~Raparthy}, \bibinfo{author}{S.~Shen}, \bibinfo{author}{S.~Wan}, \bibinfo{author}{S.~Bhosale}, \bibinfo{author}{S.~Zhang}, \bibinfo{author}{S.~Vandenhende}, \bibinfo{author}{S.~Batra}, \bibinfo{author}{S.~Whitman}, \bibinfo{author}{S.~Sootla}, \bibinfo{author}{S.~Collot}, \bibinfo{author}{S.~Gururangan}, \bibinfo{author}{S.~Borodinsky}, \bibinfo{author}{T.~Herman}, \bibinfo{author}{T.~Fowler}, \bibinfo{author}{T.~Sheasha}, \bibinfo{author}{T.~Georgiou}, \bibinfo{author}{T.~Scialom}, \bibinfo{author}{T.~Speckbacher}, \bibinfo{author}{T.~Mihaylov}, \bibinfo{author}{T.~Xiao}, \bibinfo{author}{U.~Karn}, \bibinfo{author}{V.~Goswami}, \bibinfo{author}{V.~Gupta},
  \bibinfo{author}{V.~Ramanathan}, \bibinfo{author}{V.~Kerkez}, \bibinfo{author}{V.~Gonguet}, \bibinfo{author}{V.~Do}, \bibinfo{author}{V.~Vogeti}, \bibinfo{author}{V.~Albiero}, \bibinfo{author}{V.~Petrovic}, \bibinfo{author}{W.~Chu}, \bibinfo{author}{W.~Xiong}, \bibinfo{author}{W.~Fu}, \bibinfo{author}{W.~Meers}, \bibinfo{author}{X.~Martinet}, \bibinfo{author}{X.~Wang}, \bibinfo{author}{X.~Wang}, \bibinfo{author}{X.~E. Tan}, \bibinfo{author}{X.~Xia}, \bibinfo{author}{X.~Xie}, \bibinfo{author}{X.~Jia}, \bibinfo{author}{X.~Wang}, \bibinfo{author}{Y.~Goldschlag}, \bibinfo{author}{Y.~Gaur}, \bibinfo{author}{Y.~Babaei}, \bibinfo{author}{Y.~Wen}, \bibinfo{author}{Y.~Song}, \bibinfo{author}{Y.~Zhang}, \bibinfo{author}{Y.~Li}, \bibinfo{author}{Y.~Mao}, \bibinfo{author}{Z.~D. Coudert}, \bibinfo{author}{Z.~Yan}, \bibinfo{author}{Z.~Chen}, \bibinfo{author}{Z.~Papakipos}, \bibinfo{author}{A.~Singh}, \bibinfo{author}{A.~Srivastava}, \bibinfo{author}{A.~Jain}, \bibinfo{author}{A.~Kelsey}, \bibinfo{author}{A.~Shajnfeld},
  \bibinfo{author}{A.~Gangidi}, \bibinfo{author}{A.~Victoria}, \bibinfo{author}{A.~Goldstand}, \bibinfo{author}{A.~Menon}, \bibinfo{author}{A.~Sharma}, \bibinfo{author}{A.~Boesenberg}, \bibinfo{author}{A.~Baevski}, \bibinfo{author}{A.~Feinstein}, \bibinfo{author}{A.~Kallet}, \bibinfo{author}{A.~Sangani}, \bibinfo{author}{A.~Teo}, \bibinfo{author}{A.~Yunus}, \bibinfo{author}{A.~Lupu}, \bibinfo{author}{A.~Alvarado}, \bibinfo{author}{A.~Caples}, \bibinfo{author}{A.~Gu}, \bibinfo{author}{A.~Ho}, \bibinfo{author}{A.~Poulton}, \bibinfo{author}{A.~Ryan}, \bibinfo{author}{A.~Ramchandani}, \bibinfo{author}{A.~Dong}, \bibinfo{author}{A.~Franco}, \bibinfo{author}{A.~Goyal}, \bibinfo{author}{A.~Saraf}, \bibinfo{author}{A.~Chowdhury}, \bibinfo{author}{A.~Gabriel}, \bibinfo{author}{A.~Bharambe}, \bibinfo{author}{A.~Eisenman}, \bibinfo{author}{A.~Yazdan}, \bibinfo{author}{B.~James}, \bibinfo{author}{B.~Maurer}, \bibinfo{author}{B.~Leonhardi}, \bibinfo{author}{B.~Huang}, \bibinfo{author}{B.~Loyd}, \bibinfo{author}{B.~D.
  Paola}, \bibinfo{author}{B.~Paranjape}, \bibinfo{author}{B.~Liu}, \bibinfo{author}{B.~Wu}, \bibinfo{author}{B.~Ni}, \bibinfo{author}{B.~Hancock}, \bibinfo{author}{B.~Wasti}, \bibinfo{author}{B.~Spence}, \bibinfo{author}{B.~Stojkovic}, \bibinfo{author}{B.~Gamido}, \bibinfo{author}{B.~Montalvo}, \bibinfo{author}{C.~Parker}, \bibinfo{author}{C.~Burton}, \bibinfo{author}{C.~Mejia}, \bibinfo{author}{C.~Liu}, \bibinfo{author}{C.~Wang}, \bibinfo{author}{C.~Kim}, \bibinfo{author}{C.~Zhou}, \bibinfo{author}{C.~Hu}, \bibinfo{author}{C.-H. Chu}, \bibinfo{author}{C.~Cai}, \bibinfo{author}{C.~Tindal}, \bibinfo{author}{C.~Feichtenhofer}, \bibinfo{author}{C.~Gao}, \bibinfo{author}{D.~Civin}, \bibinfo{author}{D.~Beaty}, \bibinfo{author}{D.~Kreymer}, \bibinfo{author}{D.~Li}, \bibinfo{author}{D.~Adkins}, \bibinfo{author}{D.~Xu}, \bibinfo{author}{D.~Testuggine}, \bibinfo{author}{D.~David}, \bibinfo{author}{D.~Parikh}, \bibinfo{author}{D.~Liskovich}, \bibinfo{author}{D.~Foss}, \bibinfo{author}{D.~Wang},
  \bibinfo{author}{D.~Le}, \bibinfo{author}{D.~Holland}, \bibinfo{author}{E.~Dowling}, \bibinfo{author}{E.~Jamil}, \bibinfo{author}{E.~Montgomery}, \bibinfo{author}{E.~Presani}, \bibinfo{author}{E.~Hahn}, \bibinfo{author}{E.~Wood}, \bibinfo{author}{E.-T. Le}, \bibinfo{author}{E.~Brinkman}, \bibinfo{author}{E.~Arcaute}, \bibinfo{author}{E.~Dunbar}, \bibinfo{author}{E.~Smothers}, \bibinfo{author}{F.~Sun}, \bibinfo{author}{F.~Kreuk}, \bibinfo{author}{F.~Tian}, \bibinfo{author}{F.~Kokkinos}, \bibinfo{author}{F.~Ozgenel}, \bibinfo{author}{F.~Caggioni}, \bibinfo{author}{F.~Kanayet}, \bibinfo{author}{F.~Seide}, \bibinfo{author}{G.~M. Florez}, \bibinfo{author}{G.~Schwarz}, \bibinfo{author}{G.~Badeer}, \bibinfo{author}{G.~Swee}, \bibinfo{author}{G.~Halpern}, \bibinfo{author}{G.~Herman}, \bibinfo{author}{G.~Sizov}, \bibinfo{author}{Guangyi}, \bibinfo{author}{Zhang}, \bibinfo{author}{G.~Lakshminarayanan}, \bibinfo{author}{H.~Inan}, \bibinfo{author}{H.~Shojanazeri}, \bibinfo{author}{H.~Zou}, \bibinfo{author}{H.~Wang},
  \bibinfo{author}{H.~Zha}, \bibinfo{author}{H.~Habeeb}, \bibinfo{author}{H.~Rudolph}, \bibinfo{author}{H.~Suk}, \bibinfo{author}{H.~Aspegren}, \bibinfo{author}{H.~Goldman}, \bibinfo{author}{H.~Zhan}, \bibinfo{author}{I.~Damlaj}, \bibinfo{author}{I.~Molybog}, \bibinfo{author}{I.~Tufanov}, \bibinfo{author}{I.~Leontiadis}, \bibinfo{author}{I.-E. Veliche}, \bibinfo{author}{I.~Gat}, \bibinfo{author}{J.~Weissman}, \bibinfo{author}{J.~Geboski}, \bibinfo{author}{J.~Kohli}, \bibinfo{author}{J.~Lam}, \bibinfo{author}{J.~Asher}, \bibinfo{author}{J.-B. Gaya}, \bibinfo{author}{J.~Marcus}, \bibinfo{author}{J.~Tang}, \bibinfo{author}{J.~Chan}, \bibinfo{author}{J.~Zhen}, \bibinfo{author}{J.~Reizenstein}, \bibinfo{author}{J.~Teboul}, \bibinfo{author}{J.~Zhong}, \bibinfo{author}{J.~Jin}, \bibinfo{author}{J.~Yang}, \bibinfo{author}{J.~Cummings}, \bibinfo{author}{J.~Carvill}, \bibinfo{author}{J.~Shepard}, \bibinfo{author}{J.~McPhie}, \bibinfo{author}{J.~Torres}, \bibinfo{author}{J.~Ginsburg}, \bibinfo{author}{J.~Wang},
  \bibinfo{author}{K.~Wu}, \bibinfo{author}{K.~H. U}, \bibinfo{author}{K.~Saxena}, \bibinfo{author}{K.~Khandelwal}, \bibinfo{author}{K.~Zand}, \bibinfo{author}{K.~Matosich}, \bibinfo{author}{K.~Veeraraghavan}, \bibinfo{author}{K.~Michelena}, \bibinfo{author}{K.~Li}, \bibinfo{author}{K.~Jagadeesh}, \bibinfo{author}{K.~Huang}, \bibinfo{author}{K.~Chawla}, \bibinfo{author}{K.~Huang}, \bibinfo{author}{L.~Chen}, \bibinfo{author}{L.~Garg}, \bibinfo{author}{L.~A}, \bibinfo{author}{L.~Silva}, \bibinfo{author}{L.~Bell}, \bibinfo{author}{L.~Zhang}, \bibinfo{author}{L.~Guo}, \bibinfo{author}{L.~Yu}, \bibinfo{author}{L.~Moshkovich}, \bibinfo{author}{L.~Wehrstedt}, \bibinfo{author}{M.~Khabsa}, \bibinfo{author}{M.~Avalani}, \bibinfo{author}{M.~Bhatt}, \bibinfo{author}{M.~Mankus}, \bibinfo{author}{M.~Hasson}, \bibinfo{author}{M.~Lennie}, \bibinfo{author}{M.~Reso}, \bibinfo{author}{M.~Groshev}, \bibinfo{author}{M.~Naumov}, \bibinfo{author}{M.~Lathi}, \bibinfo{author}{M.~Keneally}, \bibinfo{author}{M.~Liu},
  \bibinfo{author}{M.~L. Seltzer}, \bibinfo{author}{M.~Valko}, \bibinfo{author}{M.~Restrepo}, \bibinfo{author}{M.~Patel}, \bibinfo{author}{M.~Vyatskov}, \bibinfo{author}{M.~Samvelyan}, \bibinfo{author}{M.~Clark}, \bibinfo{author}{M.~Macey}, \bibinfo{author}{M.~Wang}, \bibinfo{author}{M.~J. Hermoso}, \bibinfo{author}{M.~Metanat}, \bibinfo{author}{M.~Rastegari}, \bibinfo{author}{M.~Bansal}, \bibinfo{author}{N.~Santhanam}, \bibinfo{author}{N.~Parks}, \bibinfo{author}{N.~White}, \bibinfo{author}{N.~Bawa}, \bibinfo{author}{N.~Singhal}, \bibinfo{author}{N.~Egebo}, \bibinfo{author}{N.~Usunier}, \bibinfo{author}{N.~Mehta}, \bibinfo{author}{N.~P. Laptev}, \bibinfo{author}{N.~Dong}, \bibinfo{author}{N.~Cheng}, \bibinfo{author}{O.~Chernoguz}, \bibinfo{author}{O.~Hart}, \bibinfo{author}{O.~Salpekar}, \bibinfo{author}{O.~Kalinli}, \bibinfo{author}{P.~Kent}, \bibinfo{author}{P.~Parekh}, \bibinfo{author}{P.~Saab}, \bibinfo{author}{P.~Balaji}, \bibinfo{author}{P.~Rittner}, \bibinfo{author}{P.~Bontrager},
  \bibinfo{author}{P.~Roux}, \bibinfo{author}{P.~Dollar}, \bibinfo{author}{P.~Zvyagina}, \bibinfo{author}{P.~Ratanchandani}, \bibinfo{author}{P.~Yuvraj}, \bibinfo{author}{Q.~Liang}, \bibinfo{author}{R.~Alao}, \bibinfo{author}{R.~Rodriguez}, \bibinfo{author}{R.~Ayub}, \bibinfo{author}{R.~Murthy}, \bibinfo{author}{R.~Nayani}, \bibinfo{author}{R.~Mitra}, \bibinfo{author}{R.~Parthasarathy}, \bibinfo{author}{R.~Li}, \bibinfo{author}{R.~Hogan}, \bibinfo{author}{R.~Battey}, \bibinfo{author}{R.~Wang}, \bibinfo{author}{R.~Howes}, \bibinfo{author}{R.~Rinott}, \bibinfo{author}{S.~Mehta}, \bibinfo{author}{S.~Siby}, \bibinfo{author}{S.~J. Bondu}, \bibinfo{author}{S.~Datta}, \bibinfo{author}{S.~Chugh}, \bibinfo{author}{S.~Hunt}, \bibinfo{author}{S.~Dhillon}, \bibinfo{author}{S.~Sidorov}, \bibinfo{author}{S.~Pan}, \bibinfo{author}{S.~Mahajan}, \bibinfo{author}{S.~Verma}, \bibinfo{author}{S.~Yamamoto}, \bibinfo{author}{S.~Ramaswamy}, \bibinfo{author}{S.~Lindsay}, \bibinfo{author}{S.~Lindsay}, \bibinfo{author}{S.~Feng},
  \bibinfo{author}{S.~Lin}, \bibinfo{author}{S.~C. Zha}, \bibinfo{author}{S.~Patil}, \bibinfo{author}{S.~Shankar}, \bibinfo{author}{S.~Zhang}, \bibinfo{author}{S.~Zhang}, \bibinfo{author}{S.~Wang}, \bibinfo{author}{S.~Agarwal}, \bibinfo{author}{S.~Sajuyigbe}, \bibinfo{author}{S.~Chintala}, \bibinfo{author}{S.~Max}, \bibinfo{author}{S.~Chen}, \bibinfo{author}{S.~Kehoe}, \bibinfo{author}{S.~Satterfield}, \bibinfo{author}{S.~Govindaprasad}, \bibinfo{author}{S.~Gupta}, \bibinfo{author}{S.~Deng}, \bibinfo{author}{S.~Cho}, \bibinfo{author}{S.~Virk}, \bibinfo{author}{S.~Subramanian}, \bibinfo{author}{S.~Choudhury}, \bibinfo{author}{S.~Goldman}, \bibinfo{author}{T.~Remez}, \bibinfo{author}{T.~Glaser}, \bibinfo{author}{T.~Best}, \bibinfo{author}{T.~Koehler}, \bibinfo{author}{T.~Robinson}, \bibinfo{author}{T.~Li}, \bibinfo{author}{T.~Zhang}, \bibinfo{author}{T.~Matthews}, \bibinfo{author}{T.~Chou}, \bibinfo{author}{T.~Shaked}, \bibinfo{author}{V.~Vontimitta}, \bibinfo{author}{V.~Ajayi}, \bibinfo{author}{V.~Montanez},
  \bibinfo{author}{V.~Mohan}, \bibinfo{author}{V.~S. Kumar}, \bibinfo{author}{V.~Mangla}, \bibinfo{author}{V.~Ionescu}, \bibinfo{author}{V.~Poenaru}, \bibinfo{author}{V.~T. Mihailescu}, \bibinfo{author}{V.~Ivanov}, \bibinfo{author}{W.~Li}, \bibinfo{author}{W.~Wang}, \bibinfo{author}{W.~Jiang}, \bibinfo{author}{W.~Bouaziz}, \bibinfo{author}{W.~Constable}, \bibinfo{author}{X.~Tang}, \bibinfo{author}{X.~Wu}, \bibinfo{author}{X.~Wang}, \bibinfo{author}{X.~Wu}, \bibinfo{author}{X.~Gao}, \bibinfo{author}{Y.~Kleinman}, \bibinfo{author}{Y.~Chen}, \bibinfo{author}{Y.~Hu}, \bibinfo{author}{Y.~Jia}, \bibinfo{author}{Y.~Qi}, \bibinfo{author}{Y.~Li}, \bibinfo{author}{Y.~Zhang}, \bibinfo{author}{Y.~Zhang}, \bibinfo{author}{Y.~Adi}, \bibinfo{author}{Y.~Nam}, \bibinfo{author}{Yu}, \bibinfo{author}{Wang}, \bibinfo{author}{Y.~Zhao}, \bibinfo{author}{Y.~Hao}, \bibinfo{author}{Y.~Qian}, \bibinfo{author}{Y.~Li}, \bibinfo{author}{Y.~He}, \bibinfo{author}{Z.~Rait}, \bibinfo{author}{Z.~DeVito}, \bibinfo{author}{Z.~Rosnbrick},
  \bibinfo{author}{Z.~Wen}, \bibinfo{author}{Z.~Yang}, \bibinfo{author}{Z.~Zhao}, \bibinfo{author}{Z.~Ma}, \bibinfo{title}{The llama 3 herd of models}, \bibinfo{year}{2024}. \URLprefix \url{https://arxiv.org/abs/2407.21783}. \href{http://arxiv.org/abs/2407.21783}{{\tt arXiv:2407.21783}}.
\bibitem[{Lin(2004)}]{lin-2004-rouge}
\bibinfo{author}{C.-Y. Lin},
\newblock \bibinfo{title}{{ROUGE}: A package for automatic evaluation of summaries},
\newblock in: \bibinfo{booktitle}{Text Summarization Branches Out}, \bibinfo{publisher}{Association for Computational Linguistics}, \bibinfo{address}{Barcelona, Spain}, \bibinfo{year}{2004}, pp. \bibinfo{pages}{74--81}. \URLprefix \url{https://aclanthology.org/W04-1013/}.
\bibitem[{Zhang et~al.(2020)Zhang, Kishore, Wu, Weinberger, and Artzi}]{zhang2020bertscoreevaluatingtextgeneration}
\bibinfo{author}{T.~Zhang}, \bibinfo{author}{V.~Kishore}, \bibinfo{author}{F.~Wu}, \bibinfo{author}{K.~Q. Weinberger}, \bibinfo{author}{Y.~Artzi}, \bibinfo{title}{Bertscore: Evaluating text generation with bert}, \bibinfo{year}{2020}. \URLprefix \url{https://arxiv.org/abs/1904.09675}. \href{http://arxiv.org/abs/1904.09675}{{\tt arXiv:1904.09675}}.
\bibitem[{Golchin and Surdeanu(2025)}]{golchin-surdeanu-2025-data}
\bibinfo{author}{S.~Golchin}, \bibinfo{author}{M.~Surdeanu},
\newblock \bibinfo{title}{Data contamination quiz: A tool to detect and estimate contamination in large language models},
\newblock \bibinfo{journal}{Transactions of the Association for Computational Linguistics} \bibinfo{volume}{13} (\bibinfo{year}{2025}) \bibinfo{pages}{809--830}. \URLprefix \url{https://aclanthology.org/2025.tacl-1.37/}. \DOIprefix\doi{10.1162/tacl.a.20}.

\end{thebibliography}

\appendix
\section{Term List}
\label{app:term-list}

Table~\ref{tab:term-list} reports the full list of 118 queer-related terms used for term-based extraction. For each term, we provide the broad category assigned in our taxonomy. A machine-readable version of the resource, including subcategories and definitions, will be released along with the project repository.

\scriptsize
\begin{longtable}{@{}p{0.22\linewidth}p{0.24\linewidth}p{0.22\linewidth}p{0.24\linewidth}@{}}
\caption{Full list of queer-related terms used for extraction.}
\label{tab:term-list} \\
\toprule
\textbf{Term} & \textbf{Category} & \textbf{Term} & \textbf{Category} \\
\midrule
\endfirsthead

\toprule
\textbf{Term} & \textbf{Category} & \textbf{Term} & \textbf{Category} \\
\midrule
\endhead

\bottomrule
\endlastfoot

werk & Slang & kiki & Slang \\
masc & Identity, Slang & femme & Identity, Slang \\
yas & Intersectional & gag & Slang \\
stunt & Intersectional & glow up & Intersectional \\
trans & Identity & queer & Identity \\
homo & Identity,  & twerk & Intersectional \\
cis & Identity & two-spirit & Identity, Intersectional \\
diva & Intersectional & gurl & Intersectional \\
fag & Slang,  & bae & Intersectional \\
straight-acting & Identity  & straight-passing & Identity,  \\
slay & Intersectional & twink & Identity, Slang \\
drag & Identity, Slang & sashay & Intersectional \\
shade & Intersectional & kween & Intersectional \\
henny & Intersectional & rainbow capitalism & Slang \\
rainbow washing & Slang & coming out & Slang \\
polycule & Identity, Slang & baby gay & Identity, Slang \\
friend of dorothy & Identity, Slang & gold star lesbian & Identity, Slang \\
lipstick lesbian & Identity, Slang & bi panic & Slang \\
aro & Identity & deadname & Identity, Slang \\
sapphic & Identity & voguing & Intersectional \\
pinkwashing & Slang & quiltbag & Identity, Intersectional \\
enbian & Identity & t4t & Identity \\
zhuzh & Intersectional & zhoosh & Intersectional \\
mogai & Intersectional & tea & Intersectional \\
bicurious & Identity, Slang & biromantic & Identity, Slang \\
closeted & Identity, Slang & fluid & Identity, Slang \\
ally & Intersectional & butch & Identity, Slang \\
dyke & Identity, Slang  & ftm & Identity, Slang \\
f2m & Identity, Slang & mtf & Identity, Slang \\
m2f & Identity, Slang & gender non-conforming & Identity \\
genderfluid & Identity, Slang & genderqueer & Identity, Slang \\
mx & Identity, Intersectional & polyamory & Identity \\
polyamorous & Identity & qtpoc & Identity \\
qpoc & Identity & stud & Identity, Slang  \\
zir & Identity & ze & Identity \\
bi & Identity & lesbo & Identity,  \\
ace & Identity & dragqween & Identity, Slang \\
cishet & Identity, Intersectional & gatekeeping & Intersectional \\
no homo & Intersectional  & questioning & Intersectional \\
slur & Intersectional & spectrum & Identity, Intersectional \\
fruity & Slang  & camp & Slang \\
wlw & Identity, Slang & mlm & Identity, Slang \\
stone butch & Identity, Slang & celesbian & Identity, Slang \\
amab & Identity, Slang & afab & Identity, Slang \\
idahobit & Slang & achillean & Identity, Slang \\
bear & Identity, Slang & bicon & Identity, Slang \\
chapstick lesbian & Identity, Slang & enby & Identity, Slang \\
gaydar & Slang & gay radar & Slang \\
iron closet & Slang & off the spectrum & Slang \\
tomboy & Identity, Slang & beard & Slang \\
straightwashing & Slang & lavender marriage & Slang \\
terf & Slang & slander & Slang \\
bestea & Slang & devour & Slang \\
ate & Slang & slap & Slang \\
goat & Slang & acearo & Identity \\
aroace & Identity & twimbo & Slang \\
malewife & Slang & girlboss & Slang \\
lesbotorium & Slang & xim & Identity \\
xe & Identity & closet & Slang \\
\end{longtable}
\normalsize
\FloatBarrier

\section{Additional Dataset Statistics}
\label{app:stats}

Figure~\ref{fig:top30-terms} shows the thirty most frequent matched terms in the final dataset. The distribution is highly skewed, with a small number of terms accounting for a large share of the matched occurrences.

\begin{figure}[!htbp]
\centering
\includegraphics[width=0.70\linewidth]{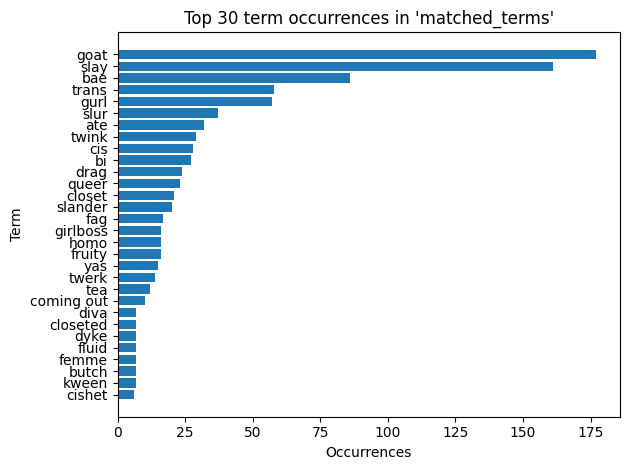}
\caption{Top 30 matched term occurrences in the final dataset.}
\label{fig:top30-terms}
\end{figure}

\FloatBarrier

Table~\ref{tab:subcategory-coverage} reports lexical coverage by taxonomy subcategory. Differently from the main categories, subcategories serve to label more specific linguistic phenomena.

\begin{table}[!htbp]
\centering
\small
\begin{tabular}{lrrrr}
\hline
\textbf{Subcategory} & \textbf{Terms} & \textbf{Matched terms} & \textbf{Coverage (\%)} & \textbf{Sentences} \\
\hline
Reclaimed & 6 & 6 & 100.0 & 71 \\
Pronoun & 5 & 4 & 80.0 & 10 \\
Shorthand & 32 & 20 & 62.5 & 182 \\
Spelling variation & 8 & 5 & 62.5 & 80 \\
Idiomatic expression & 21 & 12 & 57.1 & 254 \\
\hline
\end{tabular}
\caption{Lexical coverage and matched sentences by taxonomy subcategory. Subcategories are optional and therefore do not cover all terms in the inventory; sentence counts are non-mutually exclusive.}
\label{tab:subcategory-coverage}
\end{table}
\FloatBarrier

\section{Data Contamination Details}
\label{app:data-contamination}

We applied the Data Contamination Quiz (DCQ) framework \cite{golchin-surdeanu-2025-data} 
to assess the risk of verbatim memorisation for all models under evaluation. The method frames contamination detection as a multiple-choice task: for each dataset instance, the model is asked to identify the original sentence among four word-level perturbations of it, plus an option to select none of the provided alternatives. A consistent preference for the original instance signals prior exposure to that data.

To account for positional biases in LLMs---the tendency to favour certain 
answer positions regardless of content---the framework administers two types 
of quizzes. The Bias Detector Quiz (BDQ) identifies which answer positions the 
model systematically avoids when no correct answer is present. The Bias 
Compensator Quiz (BCQ) then places the original instance in those 
non-preferred positions and permutes it across them, allowing the framework to 
estimate a minimum and maximum contamination range while avoiding both over- 
and underestimation.

We applied DCQ to a subsample of 100 instances from our dataset. Perturbed 
alternatives are generated using GPT-5.5\footnote{\url{https://developers.openai.com/api/docs/models/gpt-5.5}} via the following prompt:

\begin{quote}
\small
\textit{You are helping create a data contamination detection quiz. Given one 
sentence, generate exactly 4 perturbed versions of it. Rules: preserve the 
original meaning and tone as closely as possible; preserve the original 
sentence structure; change only some words or short expressions; do not 
summarize, explain, or add information; do not make the sentence more concise; 
do not invent new words; only substitute with real English words or equivalent 
accredited online slang expressions; if a word or expression cannot be replaced 
naturally, keep it unchanged; each perturbation must be different from the 
others; none of the perturbations may be identical to the original sentence.}
\end{quote}

All generated perturbations were manually validated by one of the authors 
before being used in the quiz.

\section{Alternative Definitions}
\label{app:alt-definitions}

We use GPT-5.5 to generate three alternatives to the gold term definitions from our taxonomy for evaluation purposes. To do so, we use  the following prompt:

\begin{quote}
\small
\textit{You are a queer slang expert. Given a file with a column containing queer slang terms and another column containing a definition for each term, rewrite the given definition by providing 3 different alternative definitions. The new definitions must carry the same meaning as the original ones; the first alternative should be the about same length as the original one, while the next two should be progressively more concise.
Only respond with the rewritten sentence, no explanations.}
\end{quote}

All generated definitions were manually validated by the authors.

\onecolumn

\section{Manual Evaluation Examples}
\label{app:manual-evaluation}

Table~\ref{tab:manual-examples} reports selected examples from the manual
evaluation. The complete set of
manual annotations is available in the paper's repository.

\vspace{0.6em}

\begin{center}
\small
\setlength{\tabcolsep}{3pt}

\begin{tabular}{
p{0.08\textwidth}
p{0.12\textwidth}
p{0.12\textwidth}
p{0.28\textwidth}
p{0.32\textwidth}
}
\toprule
Term & Setting & Label & Output excerpt & Observation \\
\midrule

\textit{enbian}
&
Claude, Term
&
Partially 

correct
&
``A Chinese internet slang term referring to a person whose sexual
orientation or gender identity is perceived as fluid or ambiguous \ldots''
&
The output associates the term with queer identity but hallucinates a Chinese
etymology and does not capture its specific meaning.
\\

\textit{ace}
&
Claude, Term
&
Incorrect
&
``To perform something with great success or skill \ldots a person who excels
in a particular field \ldots''
&
The model defaults to common non-queer senses and fails to identify the meaning
related to the asexual spectrum.
\\

\textit{afab}
&
Llama 4, 

Term
&
Incorrect
&
``A fictional representation \ldots that shows a character to be female
\ldots''
&
The model hallucinates an unrelated meaning instead of recognizing the
established acronym \textit{assigned female at birth}.
\\

\textit{twimbo}
&
Llama 4, 

Context
&
Partially 

correct
&
``A gay man who exhibits both twink and bear-like qualities \ldots''
&
The model recognizes its association with gay male slang but incorrectly
analyzes it as a blend of \textit{twink} and \textit{bear}, rather than
\textit{twink} and \textit{bimbo}.
\\

\textit{fruity}
&
Claude, Term
&
Incorrect
&
``Having the taste, smell, or quality of fruit \ldots a voice or sound that is
rich and deep in tone.''
&
The model selects unrelated general-language senses and entirely misses the
queer slang meaning.
\\

\textit{bi panic}
&
Llama 4, 

Context
&
Partially 

correct
&
``A feeling of discomfort, alarm, or anxiety triggered by a perceived threat
to one's own bisexual identity or the bisexual community \ldots''
&
The model associates the expression with bisexuality but gives it an
excessively negative connotation, overlooking its typically lighthearted use
to describe flustered attraction.
\\

\bottomrule
\end{tabular}

\vspace{0.4em}

\refstepcounter{table}
\label{tab:manual-examples}

\begin{minipage}{0.94\textwidth}
\small
\textbf{Table \thetable.}
Selected examples from the manual evaluation. Model outputs are abbreviated
for space.
\end{minipage}
\end{center}

\vspace{1.2em}

\section*{Declaration on Generative AI}

During the preparation of this work, the author(s) used ChatGPT (OpenAI) in order to: paraphrase and reword, improve writing style, and grammar and spelling check. After using these tool(s)/service(s), the author(s) reviewed and edited the content as needed and take(s) full responsibility for the publication's content.

\end{document}